\documentclass[journal]{new-aiaa}
\usepackage[utf8]{inputenc}
\usepackage{textcomp}
\usepackage{graphicx}
\usepackage{amsmath}
\usepackage{siunitx}
\usepackage{longtable,tabularx}
\usepackage{booktabs}
\usepackage{placeins}          % provides \FloatBarrier

\title{Wind-Aware Reinforcement Learning Control of a Small Quadrotor Using Learned Onboard Wind Estimation in Simulated Atmospheric Turbulence}

\author{Abdullah Al Tasim\footnote{Graduate Research Assistant, School of Aerospace and Mechanical Engineering.}
 and Wei Sun\footnote{Corresponding Author, School of Aerospace and Mechanical Engineering; wsun@ou.edu.}}
\affil{University of Oklahoma, Norman, Oklahoma, 73019}

\begin{document}

\maketitle

\begin{abstract}
Small multirotor aircraft are increasingly tasked with operations in the atmospheric boundary layer, where turbulent winds comparable to the vehicle's airspeed degrade trajectory tracking and can defeat conventional feedback control. This work illustrates a two-stage learning pipeline that first estimates the local wind from onboard kinematics and dynamics and then exploits that estimate inside a reinforcement learning (RL) flight controller. The wind estimator, an attention-augmented gated recurrent network trained on thousands of simulated flights through von K\'arm\'an turbulence with power-law shear and veer, recovers the horizontal wind vector with a per-flight root-mean-square error of \SI{0.40}{\meter\per\second} and a direction error of \SI{3.2}{\degree} on unseen wind regimes, an accuracy near the floor imposed by unresolved turbulence, and generalizes to vertical ascent profiles with a skill score of 0.861 over a constant-wind reference. A proximal policy optimization controller receiving the frozen estimator's output reduces horizontal trajectory tracking error by 48\% relative to a wind-blind proportional-derivative baseline across mean winds of \SIrange{4}{12}{\meter\per\second}, winning on 100\% of evaluation episodes. A three-way ablation decomposes this improvement into a kinematic component, available without wind information, and a wind-perception component; the perception share rises with wind speed, from small in light winds toward roughly half the total benefit in strong winds, consistent with the quadratic scaling of aerodynamic drag. The controller degrades gracefully on out-of-distribution winds of \SIrange{13}{15}{\meter\per\second}, where the baseline fails catastrophically.
\end{abstract}

\section*{Nomenclature}

\noindent\begin{longtable*}{@{}l @{\quad=\quad} l@{}}
$U_{\mathrm{mean}}$ & reference mean wind speed at $z_{\mathrm{ref}}$, \si{\meter\per\second} \\
$z_{\mathrm{ref}}$ & reference height for the wind profile, \si{\meter} \\
$\alpha$ & power-law wind-shear exponent \\
$w_N, w_E$ & north and east components of the mean wind, \si{\meter\per\second} \\
$\hat{w}_N, \hat{w}_E$ & estimated north and east mean-wind components, \si{\meter\per\second} \\
$\tau$ & exponential-moving-average filter time constant, \si{\second} \\
$S$ & skill score relative to a constant-wind reference \\
$\mathrm{RMSE}$ & root-mean-square error \\
$\mathrm{MAE}$ & mean absolute error \\
\multicolumn{2}{@{}l}{Acronyms}\\
GRU & gated recurrent unit \\
MAE & mean absolute error \\
OOD & out-of-distribution \\
PD & proportional-derivative \\
PPO & proximal policy optimization \\
RL & reinforcement learning \\
RMSE & root-mean-square error \\
sUAS & small uncrewed aircraft system \\
\end{longtable*}

\addtocounter{table}{-1}

\section{Introduction}
Small uncrewed aircraft systems (sUAS) typically operate within the atmospheric boundary layer (ABL), where mean winds, vertical shear, and turbulent gusts routinely reach a substantial fraction of the vehicle's own airspeed and at times exceed it. For a multirotor with mass on the order of a few kilograms, the aerodynamic disturbance force scales approximately with the square of the relative wind speed, so the control burden grows sharply with wind: a disturbance that is negligible at 3~m~s$^{-1}$ becomes dominant at 12~m~s$^{-1}$. Conventional cascaded proportional-derivative (PD) and related model-based controllers treat wind purely as an unmeasured disturbance to be rejected reactively, after it has already produced a tracking error. As wind strengthens, this reactive posture leads to growing position error and, beyond the actuation margin, to divergence.

An attractive alternative is to give the controller direct knowledge of the wind. Dedicated airflow sensors add mass, cost, and calibration burden that small platforms can rarely afford, but the vehicle itself is an aerodynamic body: the mismatch between commanded thrust, measured acceleration, and observed velocity carries information about the relative wind. This motivates a two-stage architecture in which (i) a learned estimator infers the ambient wind vector from onboard kinematics and dynamics, and (ii) a learned flight controller consumes that estimate as part of its observation, enabling anticipatory rather than purely reactive disturbance handling.

This paper develops and evaluates such a pipeline in a six-degree-of-freedom simulation of a 2.59-kg quadrotor flying through von K\'arm\'an turbulence with power-law vertical shear and direction veer, physics-based sensor noise, and a library of horizontal and vertical reference trajectories. The contributions are threefold. First, it shows that an attention-augmented gated recurrent unit (GRU) network, trained on several thousand simulated flights, estimates the horizontal wind vector with a per-flight vector root-mean-square error (RMSE) of 0.40~m~s$^{-1}$ and a direction mean absolute error (MAE) of 3.2$^\circ$ on wind regimes never seen in training, an accuracy near the floor set by unresolved turbulent fluctuations, and that the same architecture extends to vertical ascent--descent flight with strong skill over a constant-wind reference. Second, it shows that a proximal policy optimization (PPO) controller conditioned on the frozen estimator's output reduces horizontal tracking RMSE by 48\% and vertical-regime horizontal-axis RMSE by 39.5\%, relative to a wind-blind PD baseline, outperforming it in every matched evaluation episode, and that a three-way ablation cleanly decomposes this improvement into a kinematic-learning component and a wind-perception component. Third, it demonstrates that the value of wind perception is regime-dependent rather than constant. In the horizontal regime the wind-perception share of the total improvement grows with mean wind speed to a mid-range peak near $U=8$~m~s$^{-1}$ before receding at the strongest winds; in the vertical regime this share rises with wind speed, from near zero in light winds toward roughly half the total benefit in strong winds. The learned controller also degrades gracefully on out-of-distribution winds, where the baseline fails outright. Together these results argue that learned wind perception is a high-leverage, sensor-free addition to small-UAS autonomy in the demanding wind conditions where it is most needed.

The remainder of the paper is organized as follows. Section~\ref{sec:litreview} reviews related work in multirotor wind estimation and learning-based flight control. Section~\ref{sec:simenv} describes the simulation environment, turbulence model, and trajectory library. Section~\ref{sec:windest} presents the wind estimation stage and its evaluation in horizontal and vertical regimes. Section~\ref{sec:rl} describes the wind-aware RL controller, baselines, and the ablation design. Section~\ref{sec:results} reports tracking results, the improvement decomposition, and the out-of-distribution probe. Section~\ref{sec:discussion} discusses mechanisms and limitations, and Section~\ref{sec:conclusion} concludes.

\section{Literature Review}\label{sec:litreview}

A hovering or slowly translating multirotor must tilt against the relative wind to hold its position, so its attitude and motor commands carry information about the wind. Model-based approaches treat the vehicle as
a calibrated drag body and invert a parametric aerodynamic model, often
within a Kalman-filtering framework. The inclination-angle method of
\cite{Neumann2015} introduced this idea, relating quadrotor pitch and
roll to the wind vector and achieving a wind-speed RMSE near
0.6~m~s$^{-1}$ after wind-tunnel calibration, while \cite{Gonzalez2019}
compared a hierarchy of kinematic, dynamic-particle, and rigid-body models
and found that the richer rigid-body model yields the most accurate
speed estimates. Such platforms have been deployed as wind sensors in
atmospheric field campaigns, where \cite{Palomaki2017} recovered
lower-atmosphere wind profiles from a hovering multirotor. A related line
augments the vehicle with dedicated airflow sensing, for example the
onboard flow sensors used by \cite{Yeo2018} for multirotor pitch control
in wind; the present work instead infers wind from existing state and
control signals alone. These model-based methods are attractive for their
transparency but degrade when the aerodynamic model is misspecified, in
aggressive maneuvers, and in strong winds where quasi-steady drag
assumptions break down.

Learning-based estimators relax these assumptions by regressing wind from
flight dynamics directly. These span a range of methods, from
nearest-neighbor regression on flight data \cite{Wang2019} to deep
residual-aerodynamics representations that capture wind effects accurately
enough to support control, most prominently the Neural-Fly line of work,
in which a learned basis of wind-dependent aerodynamic residuals enables
rapid online adaptation \cite{OConnell2022}. The present work differs in
explicitly reconstructing the wind vector itself as an interpretable
intermediate quantity, which can serve mapping and forecasting
applications in addition to control, and in quantifying estimation skill
against the floor imposed by unresolved turbulence.

Deep RL has produced quadrotor controllers that match or exceed classical
cascades in tracking accuracy and robustness, beginning with
demonstrations that neural policies can stabilize and recover a quadrotor
from arbitrary initial states \cite{Hwangbo2017}, extending to policies
that plan near-time-optimal racing trajectories \cite{Song2021} and that
outperform expert human pilots in drone racing \cite{Kaufmann2023}.
Recent evidence indicates that the advantage of learned control over
classical optimal control is most pronounced under unmodeled effects,
where a policy can discover more robust responses than a
decomposition-based controller \cite{Song2023}; wind is precisely such an
effect. Policy-gradient methods, particularly PPO
\cite{Schulman2017} as implemented in libraries such as
Stable-Baselines3 \cite{Raffin2021}, are the workhorse of this
literature. A growing line of work addresses flight under disturbances:
\cite{Huang2023} train a feedforward-feedback policy in simulation that
tracks aggressive trajectories in unsteady wind fields and augment it at
deployment with an adaptive disturbance estimator. Most such studies treat
wind either as unmodeled noise to be overcome through domain randomization
\cite{Tobin2017} or as a lumped force disturbance to be rejected.
Comparatively little work
conditions the deployed policy on an explicitly estimated wind state and
then quantifies how much of the resulting performance is attributable to
that perception channel as opposed to the policy's general kinematic
competence; this attribution question is a central focus of the present
study.

Classical adaptive and disturbance-observer-based control augments a
nominal controller with an online estimate of lumped disturbances, with
$\mathcal{L}_1$ adaptive control a representative framework offering fast
adaptation with bounded transients \cite{Hovakimyan2010}. Closer to the
multirotor setting, \cite{Bisheban2021} combine a geometric controller
with neural-network disturbance estimates to track trajectories in wind
fields. Such methods estimate the \emph{force} disturbance rather than the
wind \emph{velocity}, and their authority is typically limited to slow,
quasi-constant disturbance components. Hybrid approaches that combine
learned aerodynamic models with adaptive control blur this line, including
learned residual-force models \cite{OConnell2022} and hybrid
blade-element models such as NeuroBEM \cite{Bauersfeld2021}. The
flatness-based controller used as the classical baseline here is itself a
standard high-performance quadrotor comparator \cite{Sun2022}. The
methodology introduced here, comparing a wind-blind classical baseline, a
wind-blind learned policy, and a wind-aware learned policy under identical
replayed physics, offers a general template for attributing performance in
any such disturbance-aware architecture, independent of how the
disturbance information is obtained.
\section{Simulation Environment}\label{sec:simenv}

All experiments are conducted in a custom six-degree-of-freedom quadrotor
simulator that integrates rigid-body dynamics, a spectral turbulence
field, calibrated sensor models, and a trajectory-tracking controller.
The horizontal and vertical training datasets are produced from simulation that share identical vehicle dynamics, sensor models, wind model,
and controller, differing only in their trajectory libraries and in how
flight altitude is sampled; this shared physics guarantees that the two
estimators and the two control regimes are compared on a common
substrate.

\subsection{Vehicle and dynamics}

The simulated platform is a 2.59-kg quadrotor with diagonal inertia
tensor $J = \mathrm{diag}(0.078, 0.082, 0.14)$~kg~m$^2$ and an arm length
of 0.25~m. Each rotor produces thrust and reaction torque through
coefficients $k_T = 1.5\times10^{-5}$ and $k_M = 0.055\,k_T$, and the
four motor speeds map to collective thrust and body torques through a
fixed mixing matrix; motor dynamics are modeled as a first-order lag with
time constant 0.05~s and speeds bounded to $[0, 1200]$~rad~s$^{-1}$, with
per-axis collective thrust limited to between 0.3 and 2.5 times hover.
Translational aerodynamic drag is modeled as a quadratic force in the
body-relative wind with effective drag coefficients
$C_dA_{xy} = 0.038$~m$^2$ in the horizontal plane and
$C_dA_z = 0.10$~m$^2$ along the vertical axis at air density
$\rho = 1.225$~kg~m$^{-3}$, so the disturbance force scales with the
square of the relative wind speed. Rigid-body dynamics are integrated with
a fixed step of $dt = 2$~ms; the flatness-based controller (described
below) runs at 100~Hz ($dt_{\mathrm{ctrl}} = 10$~ms) and all signals are
logged at 10~Hz ($dt_{\mathrm{log}} = 0.10$~s), which sets the time
resolution of the estimator's input sequences.

The reference controller used during dataset generation is a flatness-based
cascade with pure PD outer position and inner
attitude loops, with position gains
$K_p = (5,5,7)$, $K_d = (3.5, 3.5, 4.5)$, attitude gains
$K_R = \mathrm{diag}(10,10,5)$ and $K_\omega = \mathrm{diag}(3.5,3.5,2)$,
a commanded-tilt limit of 35$^\circ$, and a position-error saturation of
$(12,12,15)$~m. This same wind-blind PD controller is the classical
baseline against which the reinforcement learning policies are compared in
Sections~\ref{sec:rl} and~\ref{sec:results}.

\subsection{Atmospheric wind and turbulence model}

The ambient wind is the sum of a deterministic mean profile and a
spatially correlated turbulent fluctuation. The mean horizontal wind
follows a power law in altitude,
\begin{equation}
  \bar{U}(z) = U_{\mathrm{mean}}\left(\frac{\max(z, z_{\mathrm{ref}})}{z_{\mathrm{ref}}}\right)^{\alpha},
  \label{eq:powerlaw}
\end{equation}
with reference height $z_{\mathrm{ref}} = 10$~m and shear exponent
$\alpha$, and its direction veers linearly with altitude at a rate of
3--8$^\circ$ per 100~m. The reference mean speed $U_{\mathrm{mean}}$ is
defined at $z_{\mathrm{ref}}$. The turbulent fluctuation is synthesized
from the von K\'arm\'an spectrum \cite{vonKarman1948}, the standard
continuous-turbulence model for atmospheric flight \cite{MILHDBK1797},
using the TurboGenPY spectral-mode method \cite{TURBOGENPY}
with 300 Fourier modes on a $64^3$ grid spanning a
$1200\times1200\times1200$~m domain, producing a frozen, divergence-free
turbulence cube that is sampled along the flight path. Turbulence
intensity is set per flight in the range 0.08--0.25 and scaled with the
local mean speed, with mild altitude dependence and stability-dependent
anisotropy cycled through neutral, stable, and convective regimes
\cite{Stull1988}. A configurable cap on the horizontal mean-wind
magnitude (\texttt{max\_horizontal\_wind}, 20~m~s$^{-1}$ in the
evaluation environment) clips the upper tail of
Eq.~(\ref{eq:powerlaw}); its consequence for the out-of-distribution
analysis is disclosed in Section~\ref{sec:discussion}.

Wind regimes are drawn by Latin hypercube sampling (LHS) to spread
coverage evenly over the regime parameters. The vertical-dataset sampler
spans six dimensions ($U_{\mathrm{mean}}$, direction, turbulence
intensity, shear exponent $\alpha$, veer, and a turbulence period) while
the horizontal-dataset sampler adds a seventh dimension, the fixed
flight altitude of each flight. In both cases the upper bound on $U_{\mathrm{mean}}$ is
set adaptively from the sampled shear so that the mean wind encountered at
altitude does not exceed a physical ceiling of 20~m~s$^{-1}$: the vertical
sampler bounds the wind at the maximum training altitude
($z_{\max} = 600$~m), whereas the horizontal sampler bounds it at each
flight's own fixed altitude, which lets low-altitude horizontal flights
experience strong winds without violating the ceiling. Evaluation uses a
disjoint LHS seed, so all reported estimation and control results are on
wind regimes never seen during training. The shear law in
Eq.~(\ref{eq:powerlaw}) materially constrains the usable vertical
evaluation range, an effect quantified in Section~\ref{sec:discussion}.

\subsection{Sensors}
 
Onboard measurements emulate a standard small-UAS sensor suite with
physics-based sampling rate, noise, and slowly drifting bias for each
instrument: a Global Positioning System (GPS) receiver reporting position and velocity at 10~Hz (position noise
0.7~m, velocity noise 0.15~m~s$^{-1}$, with random-walk bias), a
barometric altimeter at 25~Hz (0.8~m noise), an attitude estimate at
50~Hz (0.6$^\circ$ roll/pitch, 1.5$^\circ$ yaw), a rate gyro at 200~Hz
(0.3$^\circ$~s$^{-1}$ noise with bias drift), and a body-frame
accelerometer at 200~Hz (0.05~m~s$^{-2}$ noise with bias drift). Sensor
streams are down-sampled to the 10~Hz logging grid by zero-order hold.
Critically, the suite contains no airflow sensor of any kind: all wind
information available to the estimator must be inferred from the vehicle's
kinematic and control response.
 
The logged feature schema is the seventeen-channel ``sensor17'' set used
throughout this work. Thirteen of its channels are controller-agnostic
kinematic signals: altitude, the three velocity components, the three body
accelerations, the three Euler angles, and the three body rates. The
remaining four are controller-aware channels, namely the commanded
collective thrust $u_T$ and the three commanded body torques
$(u_{\tau_x}, u_{\tau_y}, u_{\tau_z})$, which expose the control effort the
vehicle is exerting to hold its trajectory against the wind. The
ground-truth labels logged alongside the sensor channels are the north and
east components of the mean wind, $(\mathrm{mean\_wind\_N},
\mathrm{mean\_wind\_E})$, evaluated from the turbulence-free mean profile
at the vehicle's instantaneous altitude. Because the labels exclude the
turbulent fluctuation, whose magnitude is comparable to the estimator's
residual error, the unresolved turbulence sets an irreducible floor on
achievable estimation accuracy, a point central to interpreting the
results in Section~\ref{sec:windest}.

\subsection{Trajectory library}

Two trajectory libraries are used, one per regime. The
\emph{horizontal} library holds altitude fixed within a flight (sampled
per flight over 80--550~m) and varies wind across flights through the
seventh LHS dimension; it comprises five constant-altitude patterns
chosen to expose the vehicle to wind from all relative angles within a
single flight: a loiter circle (\texttt{loiter\_circle\_h}), a racetrack
with long straight legs and 180$^\circ$ turns (\texttt{racetrack\_h}), a
figure-eight lemniscate (\texttt{figure8\_h}), a four-lobed cloverleaf
(\texttt{cloverleaf\_h}), and a lawnmower raster scan
(\texttt{lawnmower\_h}). The straight legs of the racetrack and lawnmower
give clean steady-state airspeed signatures, while the curved patterns
sweep heading continuously, providing dense coverage of wind-relative
angles for the estimator.

The \emph{vertical} library sweeps altitude continuously within each
flight so that the altitude-dependent mean-wind label varies along the
trajectory, training the estimator for altitude-conditioned inference. It
comprises four ascent profiles reaching up to 600~m: a pure vertical
climb (\texttt{vertical\_climb}), a helical spiral ascent
(\texttt{spiral\_ascent}), a three-dimensional figure-eight that sweeps
altitude while exciting horizontal motion (\texttt{figure8\_3d}), and a
lawnmower scan with a slow altitude ramp (\texttt{lawnmower\_scan}). The
\texttt{figure8\_3d} profile, which combines the largest horizontal
excitation with altitude sweep, is used for the qualitative
trajectory-overlay results in Section~\ref{sec:results}.

\begin{figure}[!htb]
\centering
\includegraphics[width=0.85\textwidth]{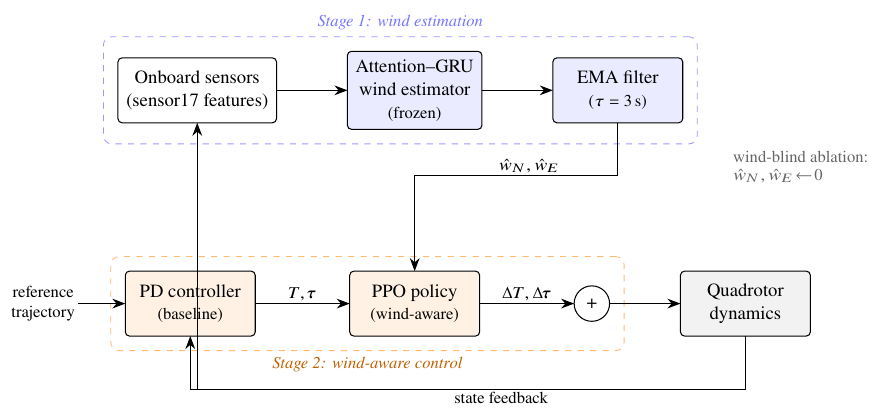}
\caption{Overview of the two-stage wind-aware control pipeline. A frozen
attention-GRU network estimates the ambient mean wind vector from the
onboard kinematic and control features; the filtered estimate augments the observation of a
PPO trajectory-tracking policy. The same wind-blind PD cascade used to
generate the training data serves as the classical baseline.}\label{fig:pipeline}
\end{figure}

\section{Wind Estimation}\label{sec:windest}

This section develops the first stage of the pipeline (Fig.~\ref{fig:pipeline}): a learned estimator
that reconstructs the mean wind from onboard signals alone.
First the methodology common to both flight regimes is described
(Section~\ref{sec:est_common}), then the horizontal estimator is presented in
full (Section~\ref{sec:est_horizontal}); the vertical estimator, which
shares the same architecture and training recipe, is presented in
Section~\ref{sec:est_vertical} as a focused description of what differs.

\subsection{Common methodology}\label{sec:est_common}

\subsubsection{Problem formulation}

The estimation task is causal sequence regression. At each logging instant
the model receives the most recent window of onboard features and predicts
the north and east components of the mean wind,
$(\mathrm{mean\_wind\_N}, \mathrm{mean\_wind\_E})$, at the vehicle's
current position. The window length is 60 samples, which at the 10~Hz
logging rate (Section~\ref{sec:simenv}) corresponds to 6~s of recent
history. Ground truth is the turbulence-free mean wind from
Eq.~(\ref{eq:powerlaw}); because the labels exclude the turbulent
fluctuation, whose instantaneous magnitude is comparable to the
estimator's residual error, the unresolved turbulence imposes an
irreducible floor on the achievable per-sample accuracy. This distinction
between per-sample (per-window) and per-flight accuracy is central to
interpreting the results and is made explicit below. The two regimes use
\emph{different} input feature representations, tailored to what is
observable in each; these are described in their respective subsections
(Sections~\ref{sec:est_horizontal} and~\ref{sec:est_vertical}).

\subsubsection{Architecture}\label{sec:est_arch}

The estimator is a recurrent network with temporal attention. A
single-layer unidirectional gated recurrent unit (GRU) \cite{Cho2014}
with 128 hidden units encodes the input window into a sequence of hidden
states. A Bahdanau-style additive attention module \cite{Bahdanau2015}
(attention dimension 128) then forms a context vector over the hidden-state
sequence, using the final hidden state as the query so that the mechanism
remains causal and the prediction at the end of the window attends over,
and only over, the samples within that window. The attention context is
concatenated with the final hidden state and passed through a three-layer
multilayer-perceptron head (Gaussian error linear unit (GELU) activations, layer normalization, dropout
0.2) that outputs the two wind components. Predictions and labels are
standardized; the training loss is a Huber (smooth-$L_1$) term on the
standardized components augmented by two small physical auxiliary terms in
unscaled space, a speed term (weight 0.03) and a direction term (weight
0.05) weighted toward higher wind speeds which improve speed calibration
and directional accuracy without distorting the dominant component
regression.

\subsubsection{Post-processing filter}\label{sec:est_filter}

Because the deployment target is a slowly varying mean wind, the
per-sample network output is smoothed by a causal exponential moving
average (EMA) before use. The filter time constant is treated as a
deployment hyperparameter; as reported per regime below, a short time
constant outperforms both the raw output and heavier smoothing, because
the network output is already accurate enough that aggressive filtering
removes genuine signal rather than noise.

\subsubsection{Metrics}\label{sec:est_metrics}
Three complementary error measures are reported, since they answer different
questions and differ substantially in magnitude.
The \emph{per-window vector RMSE} is the root-mean-square magnitude of the
two-dimensional error vector, evaluated on every sliding window in the
test set; it is the most conservative and most standard measure.
The \emph{component-mean RMSE} averages the separate north and east RMSEs
and is correspondingly smaller.
The \emph{per-flight vector RMSE} first averages the estimate over a whole
flight and then measures its error against the flight's mean-wind label;
it is the relevant measure when the deployed quantity is a per-flight
steady wind, and it is much smaller than the per-window value precisely
because per-flight averaging cancels the turbulence-driven fluctuations
that dominate the per-window error. Direction mean absolute error (MAE)
and the per-component Pearson correlation are also reported. Train,
validation, and test splits are held out by wind case (70/15/15\%), so all
reported test metrics are on wind regimes disjoint from training.

\subsection{Horizontal wind estimation}\label{sec:est_horizontal}

\subsubsection{Input features}

The horizontal estimator consumes a 48-channel feature
representation computed per flight from the
logged sensor streams. This rich representation is warranted because, in
horizontal flight at fixed altitude, the mean-wind label is nearly
constant within a flight and must be read entirely off the vehicle's body
response, with no altitude sweep to provide additional leverage. The
representation nests three groups. The first is the raw sensor17 set of
Section~\ref{sec:simenv} altitude, the three velocity and three
acceleration components, the three Euler angles, the three body rates, and
the four commanded control channels
($u_T$, $u_{\tau_x}$, $u_{\tau_y}$, $u_{\tau_z}$) together with the three
body-frame specific-force components. The second is a set of physically
motivated aerodynamic proxies: the north, east, up, and horizontal
components of the inferred aerodynamic force ($\mathrm{aero\_N}$,
$\mathrm{aero\_E}$, $\mathrm{aero\_U}$, $\mathrm{aero\_horiz}$), formed
from the residual between commanded and realized specific force, together
with the commanded-tilt magnitude and a tilt-saturation flag. These
encode the central physical cue: to hold its track against a steady wind,
the vehicle must tilt to generate a counteracting horizontal force, so the
aerodynamic residual is, to first order, a direct readout of the
wind-induced drag.

The third group is the temporal core of the representation: causal
trailing rolling means of the most informative kinematic, aerodynamic, and
control signals over two windows 10~s (100 samples, the \texttt{avg100}
channels) and 30~s (300 samples, the \texttt{avg300} channels) applied to
the horizontal velocities and accelerations, all four aerodynamic-force
components, roll and pitch, and commanded thrust. The motivation is direct:
the instantaneous aerodynamic residual is dominated by turbulent gusts,
whereas the \emph{mean} wind is revealed by the low-frequency component of
the drag signature. These trailing averages act as causal low-pass filters
that suppress the turbulent fluctuation while preserving the steady drag
bias, and they were the decisive ingredient in lowering estimation error:
earlier feature sets restricted to instantaneous signals plateaued well
above the present accuracy.

\subsubsection{Training configuration}

The horizontal estimator was trained on the merged 5-trajectory horizontal
dataset (1000 wind cases across the five constant-altitude trajectories of
Section~\ref{sec:simenv}). Sequences were built with a 60-sample window
(6~s) at a training stride of 12 samples, after a climb-aware crop that
discards each flight's initial ascent transient (it removes samples before
$\max(t_{\min}, z/\dot{z}_{\mathrm{climb}} + t_{\mathrm{safety}})$ with
$t_{\min}=8$~s), so the estimator sees only cruise-phase data where the
mean-wind label is physically meaningful. Optimization used Adam
(learning rate $3\times10^{-4}$, weight decay $10^{-5}$) with a warmup and
cosine-style decay over 25 epochs at batch size 256, with light input
noise augmentation and a wind-case-balanced sampler. Validation vector
RMSE converged to approximately 1.08~m~s$^{-1}$ with a validation
direction MAE near 3.0$^\circ$.

\subsubsection{Estimation accuracy}

On the held-out test wind cases the horizontal estimator attains a
per-window vector RMSE of 1.05~m~s$^{-1}$, a component-mean RMSE of
0.74~m~s$^{-1}$, and a direction MAE of 3.2$^\circ$, with per-component
correlations of 0.996 (north) and 0.996 (east). Averaged to the per-flight
level, the form in which the estimate is consumed by the controller, and the
vector RMSE falls to 0.40~m~s$^{-1}$. The gap between the 1.05~m~s$^{-1}$
per-window value and the 0.40~m~s$^{-1}$ per-flight value is not a
discrepancy but the signature of the turbulence floor: the per-window
error is dominated by unresolved gusts that average to near zero over a
flight, leaving a small residual mean-wind bias. Because the labels
deliberately exclude turbulence, the per-flight figure is the meaningful
measure of how well the steady wind is recovered, and at 0.40~m~s$^{-1}$
it approaches the floor set by the labeling convention itself.
Table~\ref{tab:est_h} collects these figures.

Per-trajectory results are consistent across the five patterns, with
per-flight vector RMSE ranging from 0.34~m~s$^{-1}$ on the loiter circle
to 0.51~m~s$^{-1}$ on the racetrack, and per-window direction MAE between
3.0$^\circ$ and 3.3$^\circ$ (Table~\ref{tab:est_h_traj}). The loiter
circle, whose sustained curvature continuously exposes the airframe to
wind from all relative angles, yields the cleanest per-flight estimate;
the racetrack, dominated by straight legs at two headings, is modestly
harder, consistent with richer heading coverage improving mean-wind
observability.

\begin{table}[t]
\caption{Horizontal wind estimation on held-out test wind cases. Three
RMSE conventions are reported because they differ in magnitude and meaning
(see Section~\ref{sec:est_metrics}); the per-flight value is the form
consumed by the controller.}\label{tab:est_h}
\begin{center}
\begin{tabular}{lc}
\hline\hline
Metric & Value \\
\hline
Per-window vector RMSE (m~s$^{-1}$)   & 1.05 \\
Component-mean RMSE (m~s$^{-1}$)      & 0.74 \\
Per-flight vector RMSE (m~s$^{-1}$)   & 0.40 \\
Direction MAE (deg)                  & 3.2 \\
Correlation (N, E)                   & 0.996, 0.996 \\
\hline
\end{tabular}
\end{center}
\end{table}

\begin{table}[t]
\caption{Per-trajectory horizontal estimation. Per-flight vector RMSE and
per-window direction MAE on held-out test wind cases.}\label{tab:est_h_traj}
\begin{center}
\resizebox{\columnwidth}{!}{%
\begin{tabular}{lccc}
\hline\hline
Trajectory & Per-flight RMSE (m~s$^{-1}$) & Dir.\ MAE (deg) & Test flights \\
\hline
\texttt{loiter\_circle\_h} & 0.34 & 3.0 & 18 \\
\texttt{figure8\_h}        & 0.41 & 3.1 & 20 \\
\texttt{cloverleaf\_h}     & 0.43 & 3.3 & 20 \\
\texttt{lawnmower\_h}      & 0.46 & 3.3 & 12 \\
\texttt{racetrack\_h}      & 0.51 & 3.2 & 19 \\
\hline
\end{tabular}
}
\end{center}
\end{table}

\subsubsection{Dependence on wind speed}

\begin{figure}[!htb]
\centering
\includegraphics[width=0.65\textwidth]{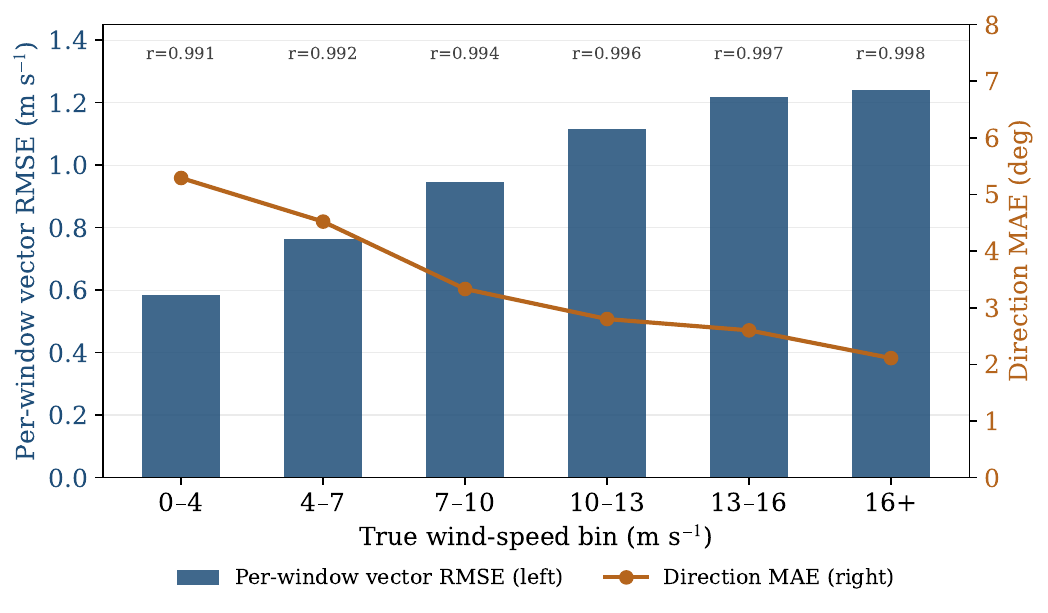}
\caption{Horizontal estimation error as a function of true wind speed.
Per-window vector RMSE (bars, left axis) rises with wind, while direction
MAE (line, right axis) falls and the per-component correlation (annotated)
increases toward 0.998. Relative accuracy therefore improves in the
strong-wind regime most relevant to control.}\label{fig:esth_speed}
\end{figure}

\begin{figure}[!htb]
\centering
\includegraphics[width=0.9\textwidth]{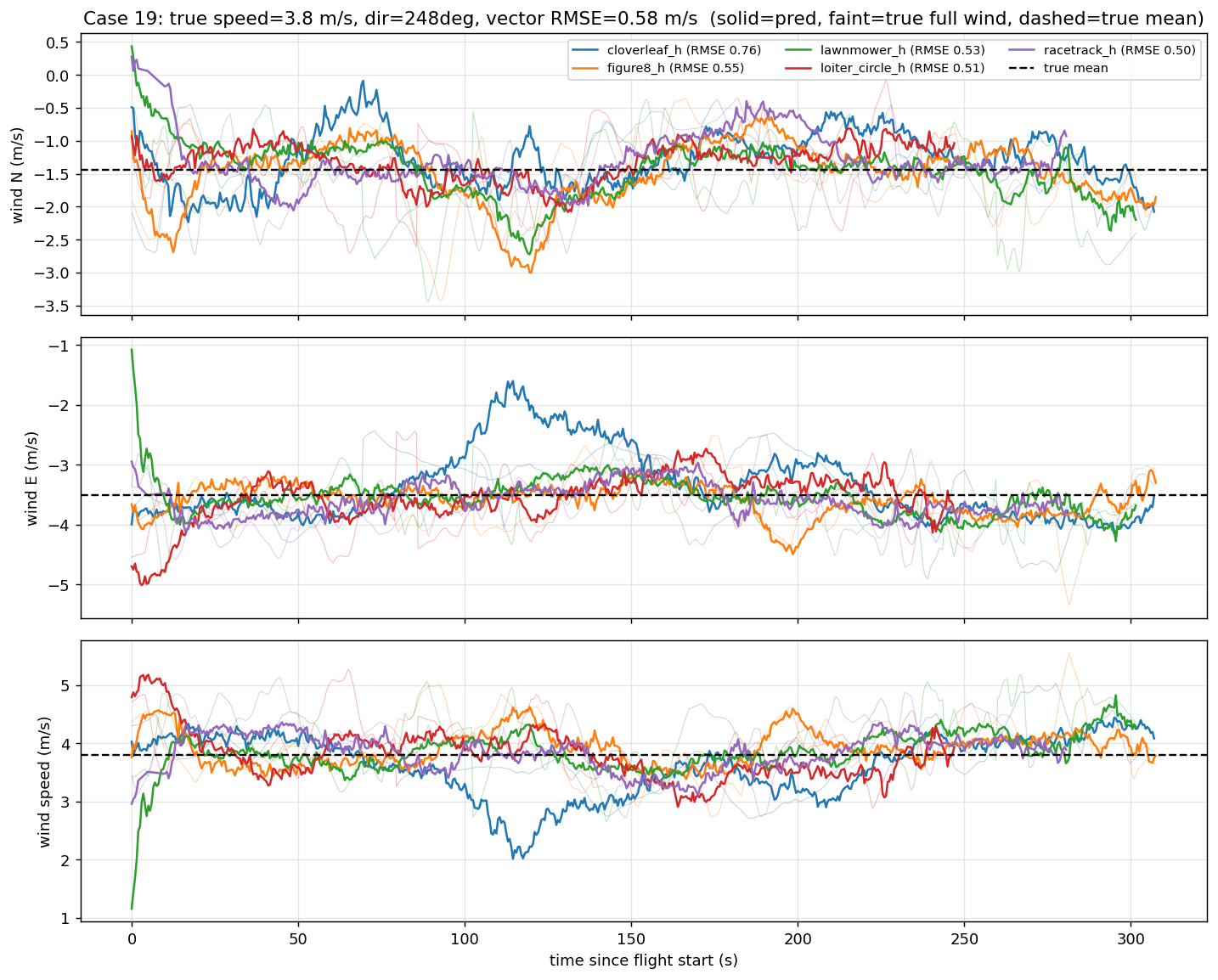}
\caption{Example horizontal estimation on a held-out low-wind test case
(true speed $\approx$3.8~m~s$^{-1}$). For each of the five trajectories,
the predicted north and east mean wind (solid) track the true mean
(dashed) through the turbulent full-wind signal (faint); the per-flight
vector RMSE ranges from 0.50 to 0.76~m~s$^{-1}$ across trajectories in
this low-wind case, the regime where relative error is largest.}\label{fig:esth_ts}
\end{figure}

The error structure varies systematically with wind speed in a way that
supports the use of the estimator for control. As the true wind increases
from the 0--4~m~s$^{-1}$ band to above 16~m~s$^{-1}$, the per-window vector
RMSE grows in absolute terms (from 0.58 to 1.24~m~s$^{-1}$), but the
direction MAE \emph{falls} monotonically (from 5.3$^\circ$ to
2.1$^\circ$) and the per-component correlation rises toward 0.998
(Fig.~\ref{fig:esth_speed}). In other words, the absolute error grows
sublinearly with wind while the \emph{relative} error shrinks: the
estimator becomes directionally sharper and more strongly correlated with
truth exactly in the strong-wind regime where wind compensation matters
most for control. This behavior follows from the drag physics and the
aerodynamic signature that the estimator reads scales with the square of
the relative wind, so the signal-to-noise ratio of the mean-wind cue
improves as wind strengthens. Figure~\ref{fig:esth_ts} shows an example
held-out low-wind test case, the regime where the relative error is
largest.

\subsection{Vertical wind estimation}\label{sec:est_vertical}

The vertical estimator shares the architecture, training loss, sequence
length, optimizer, wind-case-held-out split protocol, EMA post-filter, and
metric definitions of Section~\ref{sec:est_common}; this subsection
describes only what differs: the input representation, the training data,
and the resulting accuracy and its altitude structure.

\subsubsection{Input features}
In vertical flight the mean-wind label varies continuously along the
trajectory as the vehicle sweeps altitude through the shear and veer
profile (Section~\ref{sec:simenv}), so the estimation problem is
altitude-conditioned rather than steady-within-flight. This changes what
information is most useful, and the vertical estimator accordingly uses a
leaner 22-channel representation rather than the
horizontal postproc set. It comprises the same thirteen base kinematic
channels described in Section~\ref{sec:simenv} (altitude, the three
velocity and three acceleration components, the three Euler angles, and
the three body rates), augmented with nine reference-tracking channels:
the three position tracking errors ($\mathrm{dX}, \mathrm{dY}, \mathrm{dZ}$
between actual and commanded position) and the reference velocity and
acceleration ($\mathrm{vx\_ref}, \mathrm{vy\_ref}, \mathrm{vz\_ref}$,
$\mathrm{ax\_ref}, \mathrm{ay\_ref}, \mathrm{az\_ref}$), the latter computed
by finite differences of the known reference position along each flight.
The altitude channel itself is informative here in a way it is not for the
horizontal model: because the mean wind is a function of altitude, $z$
provides the network with a direct index into the shear profile, and the
reference-kinematic channels disambiguate commanded motion from
wind-induced deviation during the continuous climb.

\subsubsection{Training configuration and data}

The vertical estimator was trained on the merged vertical-profile dataset
of 4000 flights spanning the four
altitude-sweeping trajectories of Section~\ref{sec:simenv}
(\texttt{figure8\_3d}, \texttt{lawnmower\_scan}, \texttt{spiral\_ascent},
\texttt{vertical\_climb}) across 1000 wind cases. Sequences use the same
60-sample (6~s) window at a training stride of 16. Optimization ran for up
to 150 epochs (best validation at epoch 116) with early-stopping patience
25, otherwise matching the horizontal recipe (Adam, learning rate
$3\times10^{-4}$, batch 256, wind-case-balanced sampler), with slightly
stronger regularization (feature dropout 0.1). The model has 133{,}315
parameters. The train/validation/test split is again held out by wind case
in a 700/150/150-flight (70/15/15\%) partition.

\subsubsection{Estimation accuracy}

On the held-out test wind cases the vertical estimator attains a per-window
vector RMSE of 1.56~m~s$^{-1}$, a component-mean RMSE of 1.11~m~s$^{-1}$,
and a direction MAE of 4.9$^\circ$, with per-component correlations of
0.991 (north) and 0.991 (east) (Table~\ref{tab:est_v}). The absolute error
is larger than in the horizontal regime, which is expected: the vertical
trajectories climb to 600~m, where power-law shear amplifies the mean wind
to substantially higher magnitudes than the horizontal flights experience,
so the same fractional accuracy corresponds to a larger absolute RMSE. The
appropriate measure of the estimator's quality is therefore not the raw
RMSE but its skill against a meaningful reference.

This was quantified with a skill score $S = 1 - \mathrm{RMSE}_{\text{model}} /
\mathrm{RMSE}_{\text{ref}}$, where the reference is a constant-wind
predictor that emits the single best constant wind for the entire test set.
Because the test set pools many wind cases spanning a wide range of speeds
and directions, and because the mean wind additionally ramps with altitude
along each ascent through the shear profile, no single constant can
represent the wind well: this reference yields a vector RMSE of
11.26~m~s$^{-1}$. Against it, the estimator's 1.56~m~s$^{-1}$ corresponds
to a vector skill score of 0.861 and a speed skill score of 0.899. In other words, the estimator removes roughly 86\%
of the error that a no-altitude-knowledge constant predictor would incur,
confirming that it has learned the altitude-conditioned wind structure
rather than merely regressing to a mean. Per-trajectory results are tightly
clustered (vector RMSE 1.44--1.64~m~s$^{-1}$, direction MAE
4.8--4.9$^\circ$ across all four trajectories), indicating that accuracy
does not depend strongly on the specific maneuver.

\begin{table}[t]
\caption{Vertical wind estimation on held-out test wind cases. The skill score is computed against a constant-wind oracle that
knows the test-set mean.}\label{tab:est_v}
\begin{center}
\begin{tabular}{lc}
\hline\hline
Metric & Value \\
\hline
Per-window vector RMSE (m~s$^{-1}$)  & 1.56 \\
Component-mean RMSE (m~s$^{-1}$)     & 1.11 \\
Direction MAE (deg)                 & 4.9 \\
Correlation (N, E)                  & 0.991, 0.991 \\
Vector skill score                  & 0.861 \\
Speed skill score                   & 0.899 \\
\hline
\end{tabular}
\end{center}
\end{table}

\subsubsection{Altitude structure and post-filter}

\begin{figure}[!htb]
\centering
\includegraphics[width=0.65\textwidth]{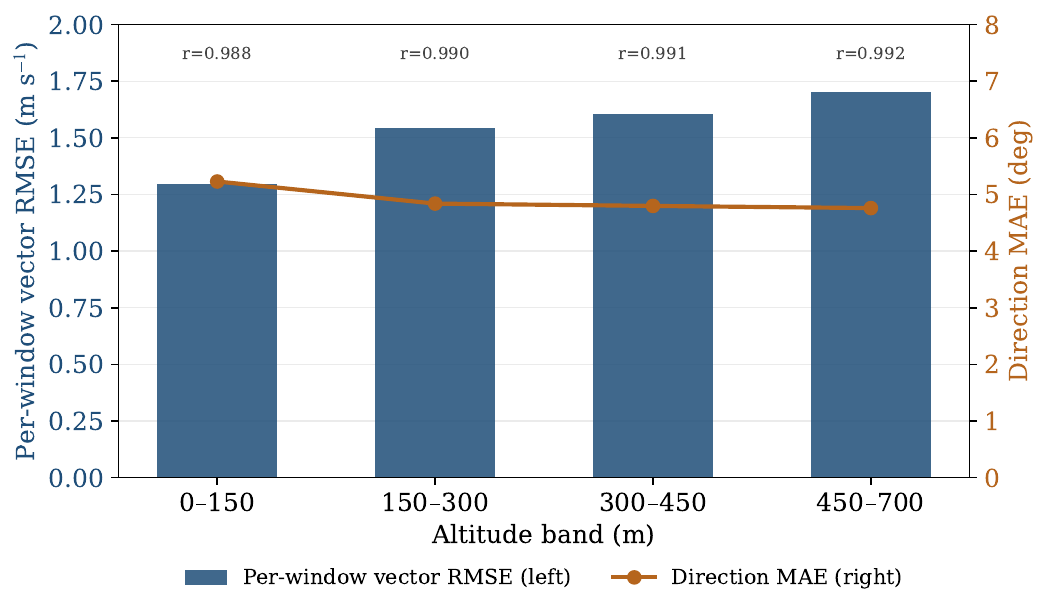}
\caption{Vertical estimation error by altitude band. Per-window vector
RMSE (bars, left axis) rises with altitude as shear amplifies the wind
magnitude, while direction MAE (line, right axis) stays near 5$^\circ$ and
the per-component correlation (annotated) remains near 0.99 throughout.
The constant directional accuracy and correlation across altitude indicate
that the error growth reflects the larger wind magnitude at altitude
rather than any loss of fractional accuracy.}\label{fig:estv_alt}
\end{figure}

Error grows gently and monotonically with altitude: per-window vector RMSE
rises from 1.29~m~s$^{-1}$ in the lowest band (0--150~m) to
1.70~m~s$^{-1}$ above 450~m, while direction MAE is essentially flat near
4.8--5.2$^\circ$ across all bands (Fig.~\ref{fig:estv_alt}). The error growth tracks the
shear-amplified increase in wind magnitude with altitude rather than any
degradation in fractional accuracy, consistent with correlation remaining
near 0.99 at every altitude. The estimator also recovers the vertical
gradient of the wind: the predicted and true altitude--speed slopes agree
closely (approximately $6.9\times10^{-3}$ vs.\ $7.2\times10^{-3}$
m~s$^{-1}$ per meter), showing that the network reproduces the shear
profile itself, not just the pointwise wind.

\begin{figure}[!htb]
\centering
\includegraphics[width=0.9\textwidth]{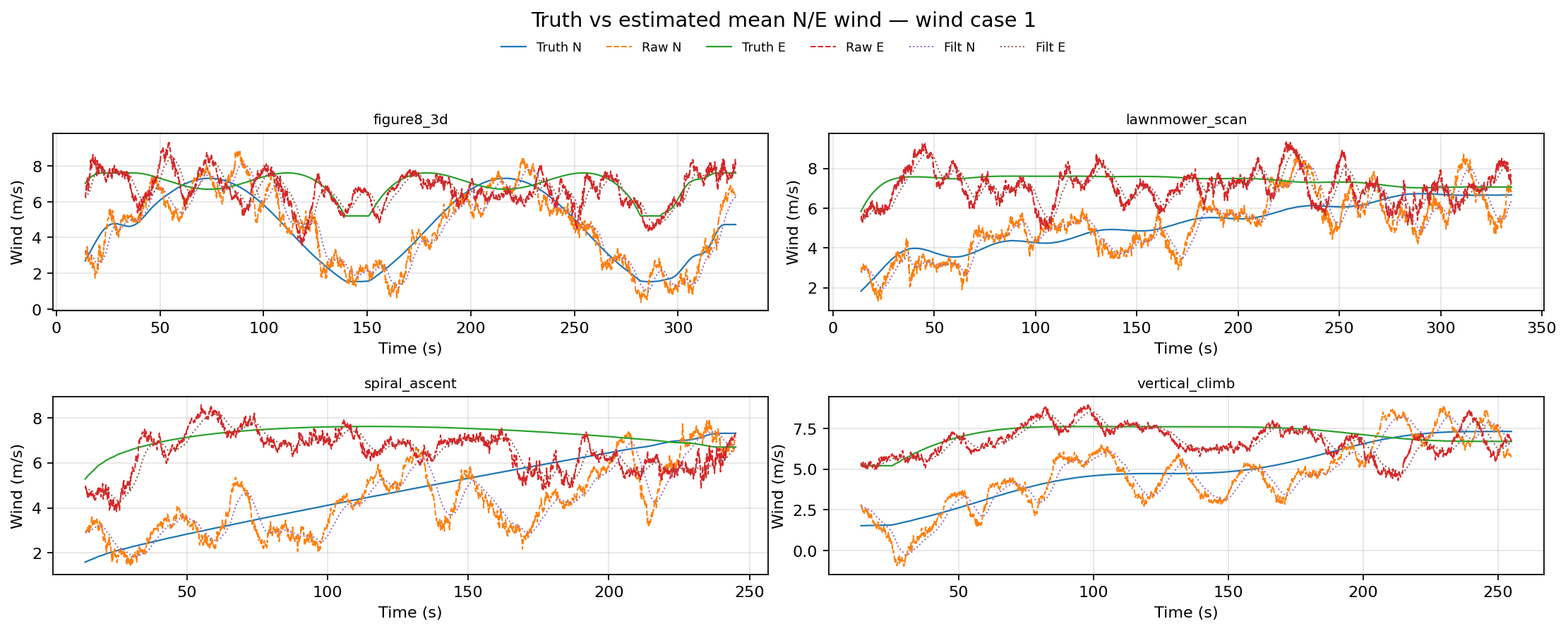}
\caption{Example vertical-regime estimation on an unseen wind case. For
each of the four ascent profiles, the true mean wind (solid) ramps with
the shear profile as the vehicle climbs, while the raw network prediction
(dashed) and the deployed $\tau=3$~s EMA-filtered estimate (dotted) track
both the altitude-driven trend and the turbulent excursions about it for
the north and east components.}\label{fig:estv_ts}
\end{figure}

The deployed EMA filter uses a short time constant of $\tau = 3$~s. On an
independently generated set of unseen wind cases (a disjoint Latin
hypercube design), the filter improves the per-window vector RMSE from
1.43~m~s$^{-1}$ raw to 1.30~m~s$^{-1}$ filtered and the direction MAE from
5.5$^\circ$ to 4.9$^\circ$; heavier smoothing was found to degrade these
metrics, because the network output is already accurate enough that a
longer time constant removes genuine signal. These unseen-case results,
slightly better than the held-out test set on raw RMSE, confirm that the
estimator generalizes to wind regimes generated entirely independently of
training (Fig.~\ref{fig:estv_ts}).

\section{Wind-Aware Reinforcement Learning Control}\label{sec:rl}

The second stage of the pipeline is a reinforcement learning (RL)
controller that consumes the frozen estimator's output and acts to reduce
trajectory-tracking error. A central design choice, which shapes both the
controller and the analysis that follows, is that the learned policy does
not replace the classical controller but \emph{augments} it: the policy
outputs a feedforward wrench correction that is added to the same
flatness-based PD controller used as the
baseline (Section~\ref{sec:simenv}). This residual-control formulation
makes the wind-blind PD law the literal zero-action limit of the
wind-aware policy, so the comparison between them is exact rather than a
comparison of two unrelated controllers.

\subsection{Policy, observation, and action}\label{sec:rl_policy}

The policy is trained with proximal policy optimization (PPO)
\cite{Schulman2017} as implemented in Stable-Baselines3
\cite{Raffin2021}, using a multilayer-perceptron actor--critic with two
hidden layers of 256 units each. Optimization uses a learning rate of
$3\times10^{-4}$, a rollout length of 2048 steps per environment, a
minibatch size of 512, 10 epochs per update, discount $\gamma = 0.99$,
generalized advantage estimation with $\lambda = 0.95$, a clip range of
0.2, an entropy coefficient of 0.005, a value-function coefficient of 0.5,
and gradient-norm clipping at 0.5. Observations are normalized by a
running mean and variance whose statistics are frozen at evaluation time.
Policies are trained for approximately 3~million environment steps across
parallel rollout environments.

\subsubsection{Observation}

The policy observes a 23-dimensional vector composed of the vehicle state,
the wind estimate, the tracking error, and the controller's own recent
output. Concretely, the observation comprises altitude; the three velocity
components; the three body-frame accelerations; the three Euler angles; the
three body rates; the two horizontal components of the estimated mean wind
$(\hat{w}_N, \hat{w}_E)$ from the frozen estimator; the three components of
the position tracking error relative to the reference; the base PD thrust
normalized by weight; and the four components of the previously applied
feedforward correction. Providing the previous feedforward and the tracking
error gives the policy the local context needed to issue a smooth,
stabilizing residual.

The wind estimate enters the observation at a single, well-defined
location, the two channels $(\hat{w}_N, \hat{w}_E)$ which is what makes
the perception ablation below clean: these two channels, and only these,
carry information the policy could not otherwise obtain. The estimator is a
frozen instance of the attention-GRU of Section~\ref{sec:windest}, loaded
from its trained checkpoint and never updated during policy training; it is
queried online at every control step and its output is passed through the
deployment EMA filter (time constant $\tau = 3$~s, with spike gating that
attenuates the filter step when an innovation exceeds three times its
running scale) before entering the observation. This preserves the
two-stage separation and ensures the controller results reflect the
estimator exactly as characterized in Section~\ref{sec:windest}.

\subsubsection{Action}

The action is a four-dimensional feedforward wrench correction,
$(\Delta T, \Delta\tau_x, \Delta\tau_y, \Delta\tau_z)$ a thrust increment
and three body-torque increments added to the base PD controller's output
at every control update. The raw policy output on $[-1, 1]^4$ is scaled by
a fixed action-scale vector $(\Delta T,\Delta\tau_{x},\Delta\tau_{y},\Delta\tau_{z})_{\max} = (10~\mathrm{N}, 5, 5, 5~\mathrm{N\,m})$,
then passed through a first-order smoother (smoothing coefficient 0.7) and a
slew-rate limiter before being applied, which prevents the policy from
issuing abrupt corrections that would excite the airframe. Because the
correction is additive on top of the PD law, a zero action recovers the
baseline controller exactly.

\subsubsection{Reward}

The per-step reward penalizes tracking error and control effort and gives a
small survival bonus:
\begin{equation}
  r = -\,\alpha\, e_{xy} \;-\; \beta\, |e_z| \;-\; \gamma\, \lVert \mathbf{u}_{\mathrm{ff}}\rVert
      \;-\; \delta\, \lVert \Delta\mathbf{u}_{\mathrm{ff}}\rVert \;+\; r_{\mathrm{alive}},
  \label{eq:reward}
\end{equation}
where $e_{xy}$ is the horizontal and $e_z$ the vertical position error,
$\mathbf{u}_{\mathrm{ff}}$ is the applied feedforward wrench, and
$\Delta\mathbf{u}_{\mathrm{ff}}$ its change since the previous step. The
horizontal-tracking weight is $\alpha = 1.0$; the vertical-tracking weight
is $\beta = 1.0$ in the horizontal regime but is reduced to $\beta = 0.5$
in the vertical regime, where trajectories span hundreds of meters of
altitude and an equal weighting would let vertical error dominate the
objective. The effort penalties use $\gamma = 0.01$ on feedforward magnitude and $\delta = 0.02$ on feedforward
slew, with a survival bonus $r_{\mathrm{alive}} = 0.05$. An episode that
violates a safety limit (ground contact, excessive tilt, or non-finite
state) is terminated with a large penalty ($-100$). All reward weights and
the action scale are saved with each run and held identical between
training and every evaluation mode.

\subsection{Training regimes}\label{sec:rl_training}

Each episode samples a fresh wind case and trajectory. In the horizontal
regime the policy trains on the five fixed-altitude trajectories of
Section~\ref{sec:simenv} at a fixed flight altitude, with mean winds
sampled over $U \in [3, 12]$~m~s$^{-1}$ and turbulence intensity over
$[0.05, 0.25]$. In the vertical regime the
policy trains on the four altitude-sweeping trajectories with an
arm-then-climb episode structure (the vehicle is pinned at the origin,
arms at $t = 1$~s, and flies the full climb to 600~m), and each episode
samples a mean wind $U_{\mathrm{mean}} \in [2, 10]$~m~s$^{-1}$, a
direction, turbulence intensity in $[0.05, 0.25]$, a shear exponent
$\alpha \in [0.08, 0.15]$, and a veer of $\pm 5^\circ$ per 100~m; the
sampled $U_{\mathrm{mean}}$ and $\alpha$ are coupled so that the shear-amplified wind at the top of the climb stays
within the estimator's training envelope (Section~\ref{sec:simenv}). The
vertical policy additionally uses a four-stage difficulty curriculum that
introduces trajectories in order of increasing complexity
(\texttt{vertical\_climb} first, then \texttt{spiral\_ascent},
\texttt{figure8\_3d}, and \texttt{lawnmower\_scan}) as training progresses.

\subsection{Baseline and the perception ablation}\label{sec:rl_ablation}

\subsubsection{Baseline}

The classical baseline is the wind-blind flatness-based PD controller of
Section~\ref{sec:simenv}, evaluated with zero feedforward correction. It is
representative of the disturbance-rejection posture of standard sUAS
autopilots: it receives no wind information and reacts to wind only through
the tracking error the wind induces. Because the RL action is additive on
top of this controller, the baseline is exactly the wind-aware policy with
its action set to zero.

\subsubsection{Three-way decomposition}

To attribute the wind-aware controller's improvement three
controllers under identical replayed physics, identical wind fields,
turbulence realizations, sensor-noise streams, trajectories, and seeds are compared via
the saved environment configuration:
(i) the wind-blind PD baseline;
(ii) the trained policy run with its wind-estimate observation channels
forced to zero (\emph{wind-blind RL}); and
(iii) the same trained policy with the wind-estimate channels active
(\emph{wind-aware RL}).
The improvement of (ii) over (i) measures what the learned residual policy
contributes from vehicle state and tracking error alone, without any wind
information, the \emph{kinematic} component. The further improvement of
(iii) over (ii) measures what the policy gains from the wind-estimate
channels, the \emph{wind-perception} component.

A methodological point deserves emphasis. The wind-blind RL controller in
(ii) is not a separately retrained policy; it is the \emph{same} trained
network evaluated with its two wind channels zeroed. The decomposition
therefore measures the inference-time value of the wind signal to the
deployed policy, holding the learned weights fixed, rather than the value
of training a policy with versus without wind. This is the appropriate and
conservative quantity for characterizing the deployed system: it isolates
exactly the information delivered through the estimator, and it cannot
benefit from any incidental differences between two separate training runs.
The decomposition characterizes the value of the
estimator's output as actually delivered, not the value of perfect wind
knowledge; because no true-wind oracle is run, no claim of estimator
sufficiency is made (Section~\ref{sec:discussion}).

\subsection{Evaluation protocol}\label{sec:rl_eval}

All controllers are evaluated on a fixed factorial sweep of held-out wind
cases generated with a sampling seed disjoint from training. The
horizontal regime comprises $N=225$ episodes (five trajectories $\times$
five mean-wind levels $\times$ three turbulence intensities $\times$ three
seeds) and the vertical regime $N=144$ episodes (four trajectories
$\times$ four mean-wind levels $\times$ three turbulence intensities
$\times$ three seeds). Every learned-policy episode is paired with a
baseline episode generated under identical replayed wind, sensor-noise,
and seed realizations, so the controllers are compared on matched
conditions and observed differences reflect the controllers alone rather
than variation in the wind or noise.

Position-tracking RMSE is recorded per episode. Aggregate improvement is
reported as the reduction in mean RMSE,
$1 - \overline{\mathrm{RMSE}}_{\mathrm{RL}}/\overline{\mathrm{RMSE}}_{\mathrm{PD}}$,
with a bootstrap 95\% confidence interval, and per-episode quantities are
reported as mean $\pm$ standard deviation. The \emph{win rate} is defined as the
fraction of matched episode pairs in which the wind-aware controller
achieves strictly lower RMSE than the baseline. To probe generalization
beyond the training envelope, an out-of-distribution (OOD) evaluation
applies the horizontal policy, unmodified, to mean winds of
13--15~m~s$^{-1}$, above its $[3,12]$~m~s$^{-1}$ training range.

\section{Results}\label{sec:results}

\subsection{Horizontal trajectory tracking}

Across the in-distribution evaluation the wind-aware RL controller reduces
mean horizontal position-tracking RMSE from $1.32 \pm 1.22$~m to
$0.68 \pm 0.23$~m, an aggregate reduction of 48.0\% (bootstrap 95\% CI
43.3--52.4\%). The wind-aware controller achieves lower error than the
baseline in all 225 matched episode pairs, and the accompanying drop in standard deviation,
from 1.22~m to 0.23~m, shows that the gain in accuracy comes with
substantially more consistent tracking (Table~\ref{tab:stats}). Evaluation
spans $U \in \{4, 6, 8, 10, 12\}$~m~s$^{-1}$ and turbulence intensities
$\{0.05, 0.15, 0.25\}$ on wind cases disjoint from training.

\begin{table}[t]
\caption{Tracking performance and paired statistics for both regimes.
RMSE values are mean $\pm$ standard deviation over $N$ matched episode
pairs. Improvement is the reduction in mean RMSE with a bootstrap 95\%
confidence interval; win rate is the fraction of matched pairs in which the wind-aware
controller achieves lower RMSE.}
\label{tab:stats}
\begin{center}
\begin{tabular}{lccccc}
\hline\hline
Regime & $N$ & PD RMSE (m) & Wind-aware (m) & Improvement (95\% CI) & Win rate \\
\hline
Horizontal      & 225 & $1.32 \pm 1.22$ & $0.68 \pm 0.23$ & 48.0\% (43.3--52.4) & 225/225 \\
Vertical ($xy$) & 144 & $0.73 \pm 0.19$ & $0.44 \pm 0.07$ & 39.5\% (37.2--41.6) & 144/144 \\
Vertical ($z$)  & 144 & $0.49 \pm 0.21$ & $0.37 \pm 0.25$ & 25.1\% (22.0--28.6) & 141/144 \\
\hline
\end{tabular}
\end{center}
\end{table}

\begin{figure}[!htb]
\centering
\includegraphics[width=0.9\textwidth]{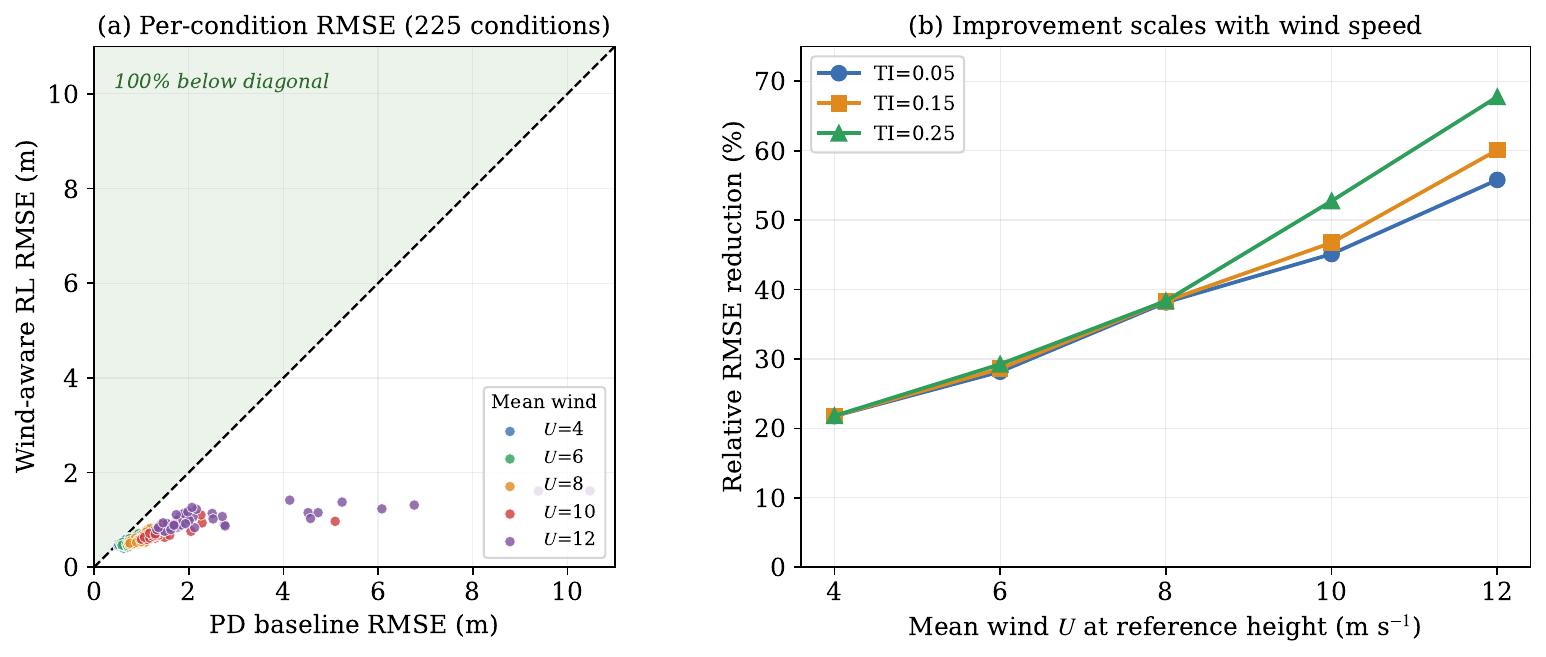}
\caption{Horizontal tracking performance across all 225 evaluation
conditions. (a)~Per-condition comparison of wind-aware RL against the PD
baseline under identical replayed physics; every condition lies below the
diagonal, so the wind-aware controller wins in all 225 cases, with the
margin widening at the strongest winds (rightmost points). (b)~Relative
RMSE reduction as a function of mean wind speed for the three turbulence
intensities. The improvement grows monotonically with wind, from
$\approx$22\% at $U=4$~m~s$^{-1}$ to 56--68\% at $U=12$~m~s$^{-1}$, and is
nearly independent of turbulence intensity at low wind, separating only at
high wind where the strongest, most turbulent cases benefit
most.}\label{fig:rlpd_h}
\end{figure}

This dominance is uniform across conditions: Figure~\ref{fig:rlpd_h}(a)
shows all 225 per-condition points lying below the diagonal, and
Figure~\ref{fig:rlpd_h}(b) shows the relative improvement growing
monotonically with wind speed, from roughly 22\% at $U=4$~m~s$^{-1}$ to
between 56\% and 68\% at $U=12$~m~s$^{-1}$ depending on turbulence
intensity.

Table~\ref{tab:hres} reports the per-wind-level breakdown. The advantage
widens sharply with wind speed: the improvement over the baseline grows
from 21.8\% at $U = 4$~m~s$^{-1}$ to 62.4\% at $U = 12$~m~s$^{-1}$, because
the baseline degrades steeply as the disturbance grows (its mean RMSE rises
from 0.65~m to 2.74~m) while the wind-aware controller degrades only
gradually (0.51~m to 1.03~m). Improvement is consistent across all five
trajectories, ranging from 33.0\% on the cloverleaf to 59.0\% on the
lawnmower (Table~\ref{tab:hres_traj}); the largest gains occur on the
trajectories with the longest sustained crosswind legs, where anticipatory
compensation has the most room to act.

\begin{figure}[!htb]
\centering
\includegraphics[width=0.9\textwidth]{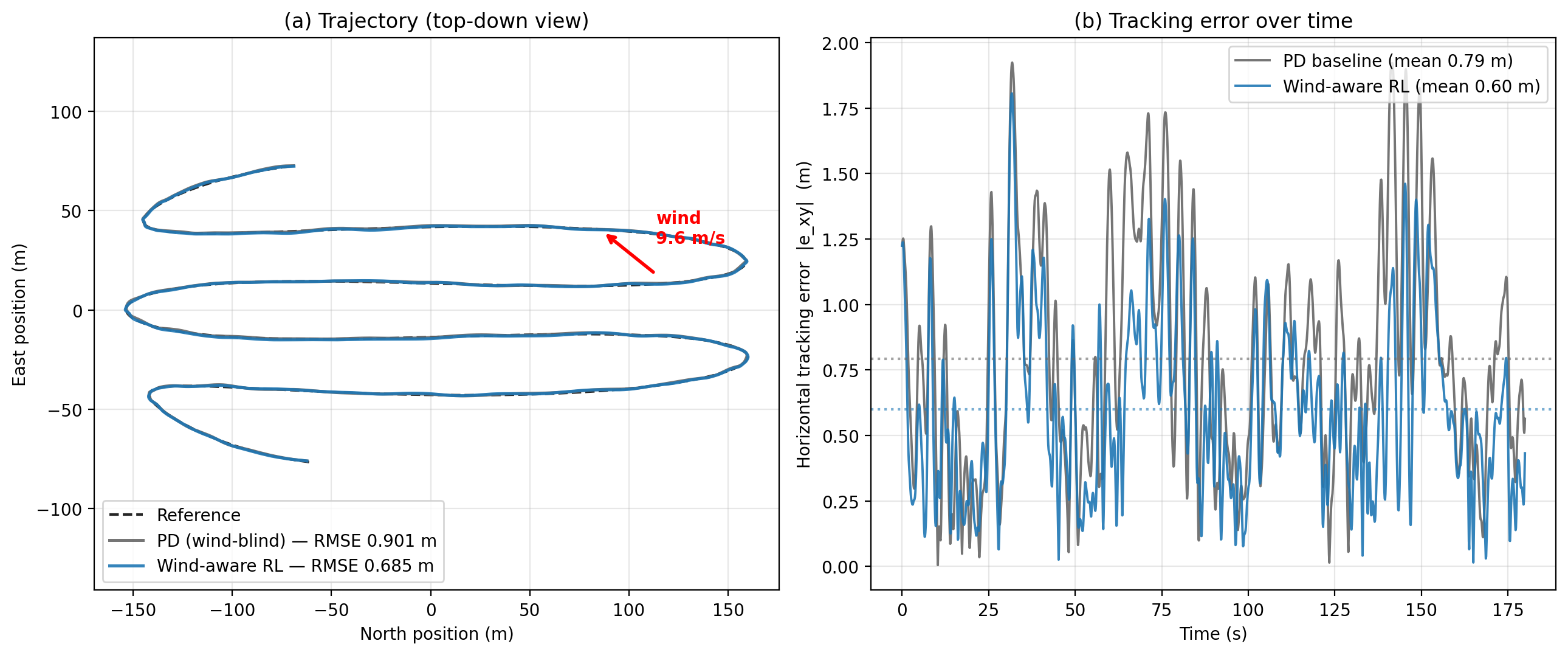}
\caption{Example horizontal tracking on the lawnmower survey pattern
(reference wind $U_{\mathrm{mean}}=6$~m~s$^{-1}$ at 10~m; encountered wind
$\approx$9.6~m~s$^{-1}$ after power-law shear amplification at the flight
altitude), evaluated under identical replayed wind and noise.
(a)~Top-down view: the wind-blind PD baseline and the wind-aware RL
controller both follow the reference (dashed), with the PD path bowing
away on the long crosswind legs. (b)~Horizontal tracking error over time:
the wind-aware policy holds a lower error for most of the flight, reducing
the mean from 0.79~m to 0.60~m and the episode RMSE from 0.901~m to
0.685~m (a 23.9\% reduction).}\label{fig:traceh}
\end{figure}

The qualitative character of the improvement is shown in
Fig.~\ref{fig:traceh}: at $U = 12$~m~s$^{-1}$ on the lawnmower survey
pattern, the PD baseline is blown persistently off-track on the crosswind
legs, while the wind-aware policy holds the reference by issuing a
feedforward correction that anticipates the steady wind.

\begin{table}[t]
\caption{Horizontal tracking by mean wind speed: mean position RMSE for the
PD baseline, wind-blind RL (policy with wind channels zeroed), and
wind-aware RL, with the improvement over baseline and the wind-perception
share of the total improvement. Each row averages 45 episodes.}
\label{tab:hres}
\begin{center}
\begin{tabular}{cccccc}
\hline\hline
$U$ & PD & Wind-blind & Wind-aware & Impr. & Perception \\
(m~s$^{-1}$) & RMSE (m) & RL (m) & RL (m) & (\%) & share (\%) \\
\hline
4  & 0.650 & 0.562 & 0.508 & 21.8 & 38.0 \\
6  & 0.745 & 0.663 & 0.531 & 28.7 & 61.7 \\
8  & 0.962 & 0.836 & 0.594 & 38.3 & 65.9 \\
10 & 1.477 & 1.182 & 0.758 & 48.7 & 59.0 \\
12 & 2.743 & 1.679 & 1.031 & 62.4 & 37.8 \\
\hline
Agg. & 1.315 & 0.985 & 0.684 & 48.0 & 47.6 \\
\hline
\end{tabular}
\end{center}
\end{table}

\begin{table}[t]
\caption{Horizontal tracking by trajectory: mean PD-baseline and
wind-aware RMSE and the improvement, aggregated over all wind levels,
turbulence intensities, and seeds (45 episodes each).}\label{tab:hres_traj}
\begin{center}
\resizebox{\columnwidth}{!}{%
\begin{tabular}{lccc}
\hline\hline
Trajectory & PD RMSE (m) & Wind-aware (m) & Impr.\ (\%) \\
\hline
\texttt{cloverleaf\_h}     & 0.947 & 0.635 & 33.0 \\
\texttt{racetrack\_h}      & 1.232 & 0.732 & 40.6 \\
\texttt{figure8\_h}        & 1.079 & 0.592 & 45.1 \\
\texttt{loiter\_circle\_h} & 1.327 & 0.646 & 51.3 \\
\texttt{lawnmower\_h}      & 1.992 & 0.817 & 59.0 \\
\hline
\end{tabular}
}
\end{center}
\end{table}

\subsection{Decomposition of the improvement}

The three-way ablation attributes the 48.0\% aggregate improvement to two
additive components under identical replayed physics. Zeroing the wind
channels of the trained policy (wind-blind RL) recovers a mean RMSE of
0.985~m, versus 1.315~m for the PD baseline and 0.684~m for the full
wind-aware policy. The step from baseline to wind-blind RL, a 25.1\%
reduction, is the \emph{kinematic} component: what the learned residual
policy contributes from vehicle state and tracking error alone. The further
step from wind-blind to wind-aware RL, a 30.5\% reduction relative to
wind-blind RL, is the \emph{wind-perception} component: what the policy gains
from the estimated wind. In terms of shares of the total
0.631~m improvement, kinematics accounts for 52.4\% and wind perception for
47.6\%, so nearly half of the benefit is attributable specifically to
sensing the wind.

\begin{figure}[!htb]
\centering
\includegraphics[width=\textwidth]{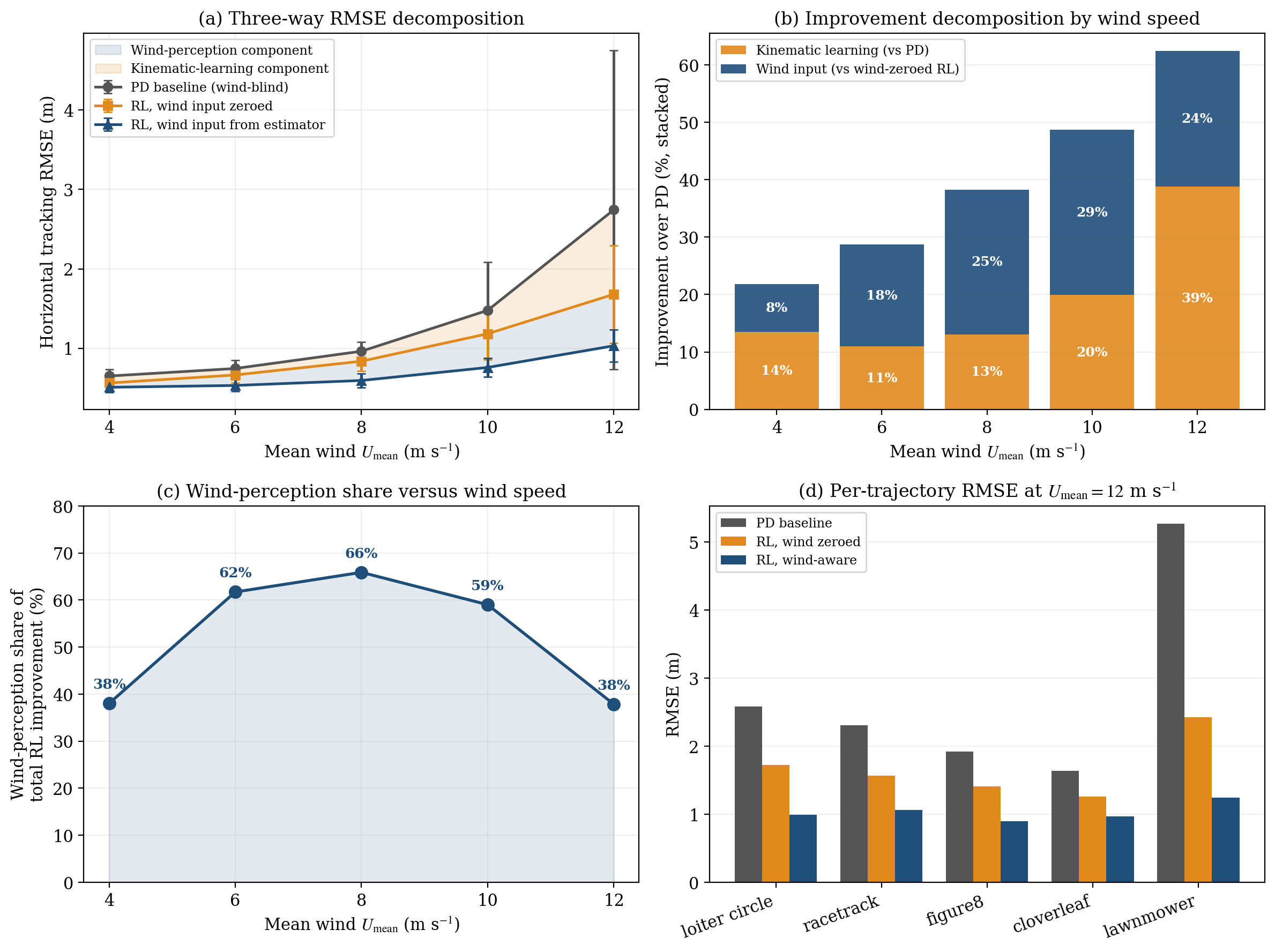}
\caption{Three-way decomposition of the horizontal tracking improvement
into kinematic and wind-perception components, under identical replayed
physics. Top left: mean tracking RMSE for the PD baseline, the wind-blind
RL policy (wind channels forced to zero), and the wind-aware RL policy
versus mean wind speed; the shaded gap between PD and wind-blind RL is the
kinematic component, and the further gap to wind-aware RL is the
wind-perception component, both widening sharply with wind. Top right: the
same improvement over PD expressed as stacked percentage contributions per
wind level, with the wind-input share growing from 8\% at
$U=4$~m~s$^{-1}$ to a mid-range maximum of 29\% at $U=10$~m~s$^{-1}$. Bottom left: wind-perception
share of the total RL improvement, which rises from 38\% at
$U=4$~m~s$^{-1}$ to a mid-range peak of 66\% near $U=8$~m~s$^{-1}$ before
receding to 38\% at $U=12$~m~s$^{-1}$, as gross reactive disturbance
rejection (available to the wind-blind policy) accounts for the largest
single increment over the baseline at the strongest winds. Bottom right:
per-trajectory RMSE at $U=12$~m~s$^{-1}$ for the three controllers,
showing the ordering PD $>$ wind-blind RL $>$ wind-aware RL on every
trajectory, most dramatically on the lawnmower
pattern.}\label{fig:decomph}
\end{figure}

The decomposition's dependence on wind speed (Fig.~\ref{fig:decomph}) is
revealing. The wind-perception share is smallest at the extremes of the
range and largest in the middle, rising from 38.0\% at $U = 4$~m~s$^{-1}$
to a peak of 65.9\% at $U = 8$~m~s$^{-1}$ before falling back to 37.8\% at
$U = 12$~m~s$^{-1}$. The rise is the expected effect: as wind strengthens,
the quadratic growth of aerodynamic drag makes anticipatory compensation
from the wind estimate increasingly valuable relative to purely reactive
kinematic control. The fall at the highest wind is also interpretable: at
$U = 12$~m~s$^{-1}$ the disturbance becomes large enough that even
wind-blind RL must devote most of its residual authority to gross
disturbance rejection, and the largest single increment over the baseline
comes from that kinematic correction, so the kinematic share grows again.
The wind-perception contribution is thus substantial across the whole
range and dominant in the mid-range, while never being the sole source of
improvement consistent with the additive residual-control design.

\subsection{Vertical trajectory tracking}
 
\begin{figure}[!htb]
\centering
\includegraphics[width=0.9\textwidth]{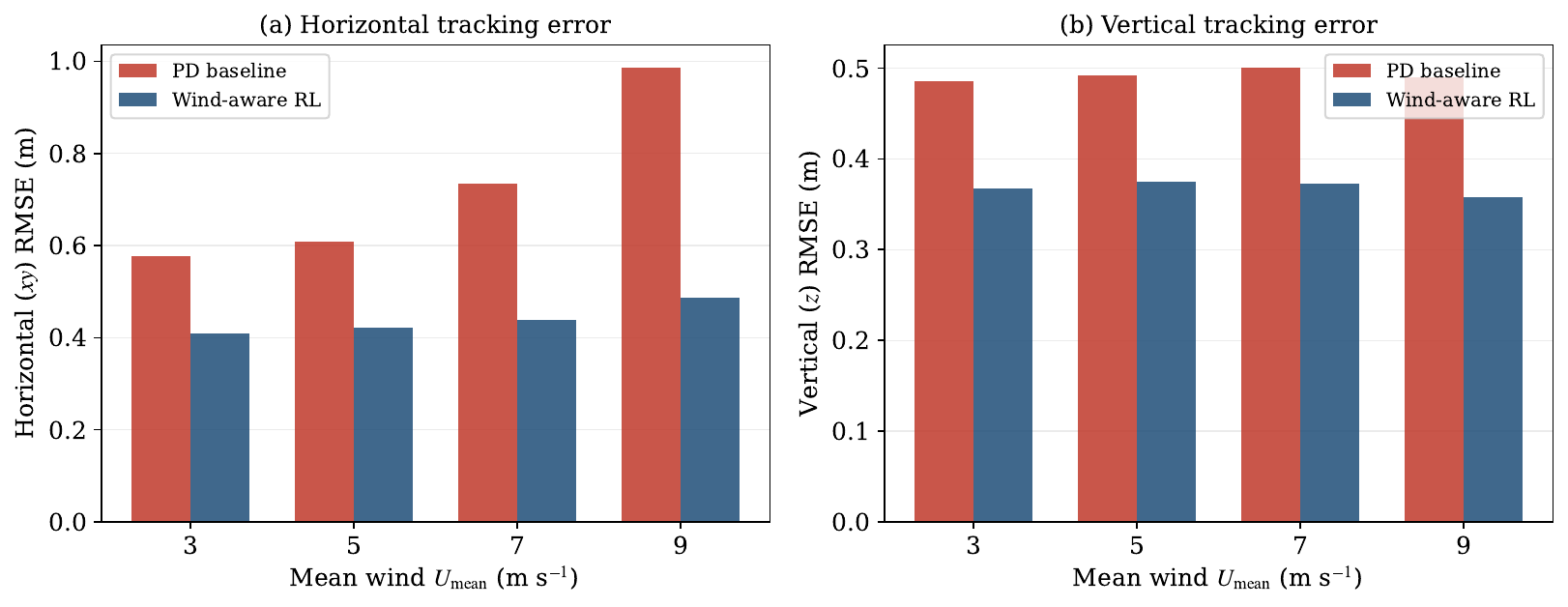}
\caption{Vertical-regime tracking comparison between the wind-blind PD
baseline and the wind-aware RL controller by mean wind speed.
(a)~Horizontal ($xy$) RMSE: the wind-aware advantage widens with wind, as
the estimated horizontal wind enables anticipatory compensation
(aggregate 0.726~m versus 0.440~m, a 39.5\% reduction). (b)~Vertical ($z$)
RMSE: the wind-aware policy improves on the baseline by a roughly constant
margin independent of horizontal wind speed (aggregate 0.492~m versus
0.369~m, a 25.1\% reduction), consistent with the $z$-axis benefit arising
from kinematic thrust management rather than wind
perception.}\label{fig:rlpdv}
\end{figure}

In the vertical regime the policy was evaluated on 144 held-out episodes
(four ascent trajectories $\times$ four mean-wind levels
$U_{\mathrm{mean}} \in \{3,5,7,9\}$~m~s$^{-1}$ $\times$ three turbulence
intensities $\times$ three seeds), with shear amplifying the wind at
altitude. Because the estimated wind is horizontal, horizontal ($xy$) and
vertical ($z$) tracking error are reported separately. The wind-aware
policy reduces mean horizontal RMSE from $0.726 \pm 0.187$~m to
$0.440 \pm 0.068$~m (39.5\% reduction, 95\% CI 37.2--41.6\%) and mean
vertical RMSE from
$0.492 \pm 0.214$~m to $0.369 \pm 0.254$~m (25.1\% reduction, 95\% CI
22.0--28.6\%); the combined position RMSE
($\sqrt{\mathrm{RMSE}_{xy}^2 + \mathrm{RMSE}_z^2}$) falls by 34.2\%, from
$0.899 \pm 0.204$~m to $0.592 \pm 0.218$~m (Fig.~\ref{fig:rlpdv},
Table~\ref{tab:stats}). The wind-aware policy achieves lower horizontal
RMSE than the baseline in all 144 matched episode pairs and lower combined
RMSE in 143 of 144. The improvement reflects both the reduced mean error and the markedly
lower variance of the wind-aware controller. Table~\ref{tab:vres} gives the per-wind-level breakdown of the
horizontal error: improvement grows with wind from 28.9\% at
$U_{\mathrm{mean}} = 3$~m~s$^{-1}$ to 50.6\% at 9~m~s$^{-1}$, mirroring the
horizontal-regime trend.
 
\begin{figure}[!htb]
\centering
\includegraphics[width=0.65\textwidth]{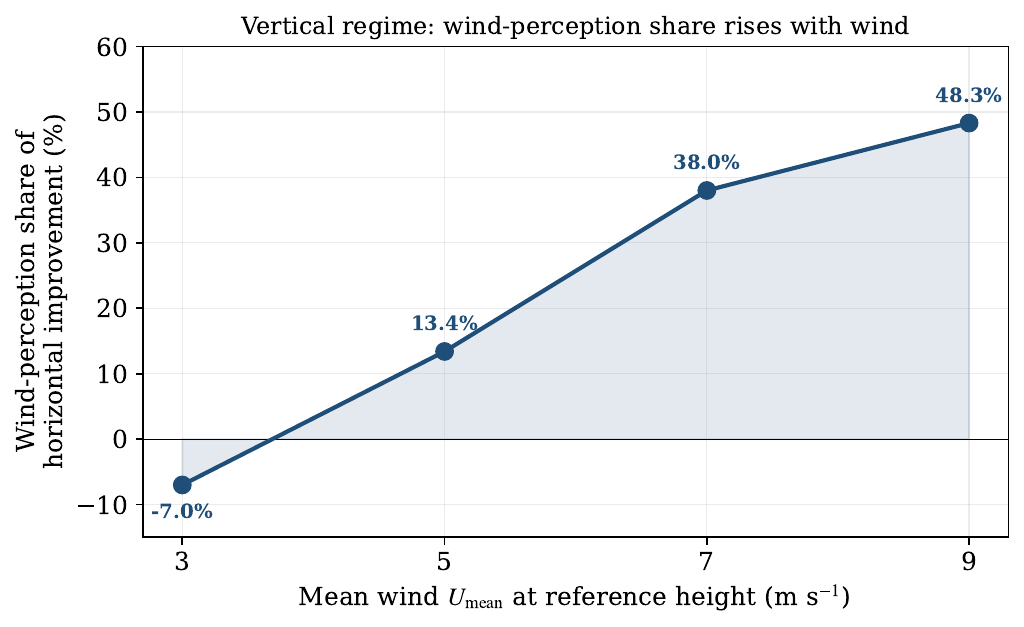}
\caption{Decomposition of the vertical-regime horizontal-tracking
improvement: wind-perception share of the total improvement as a function
of mean wind speed. The share rises monotonically from $-7.0$\% at
$U_{\mathrm{mean}}=3$~m~s$^{-1}$, where the wind channels act as slight
observation noise, through 13.4\% and 38.0\% to 48.3\% at
$U_{\mathrm{mean}}=9$~m~s$^{-1}$, crossing from counterproductive to
beneficial between 3 and 5~m~s$^{-1}$.}\label{fig:decompv}
\end{figure}

The decomposition separates cleanly along the two axes and is the central
result of this regime. For horizontal tracking, where the estimated wind
is informative, the wind-perception share of the improvement rises
monotonically with wind speed: it is $-7.0$\% at
$U_{\mathrm{mean}} = 3$~m~s$^{-1}$, where at low wind the wind channels are
very slightly counterproductive, acting as observation noise when the
disturbance is negligible, then climbs through 13.4\% and 38.0\% to 48.3\%
at $U_{\mathrm{mean}} = 9$~m~s$^{-1}$ (Fig.~\ref{fig:decompv}). This is a
cleaner monotonic trend than the horizontal regime showed, and it directly
supports the central hypothesis: wind perception becomes more valuable
precisely as the wind-induced disturbance grows. For vertical-axis
tracking, by contrast, the decomposition attributes essentially all of the
improvement to the kinematic component (104.7\%, with a wind-perception
contribution of $-4.7$\%): the estimated horizontal wind carries little
information relevant to altitude control, so the $z$-axis advantage of the
RL policy comes almost entirely from better thrust management, and
supplying the wind estimate is marginally counterproductive there. This
double dissociation, in which wind perception helps horizontal tracking
and only horizontal tracking, is strong internal evidence that the
perception channel acts through the physically correct mechanism rather
than as a generic capacity boost.
 
\begin{figure}[!htb]
\centering
\includegraphics[width=0.65\textwidth]{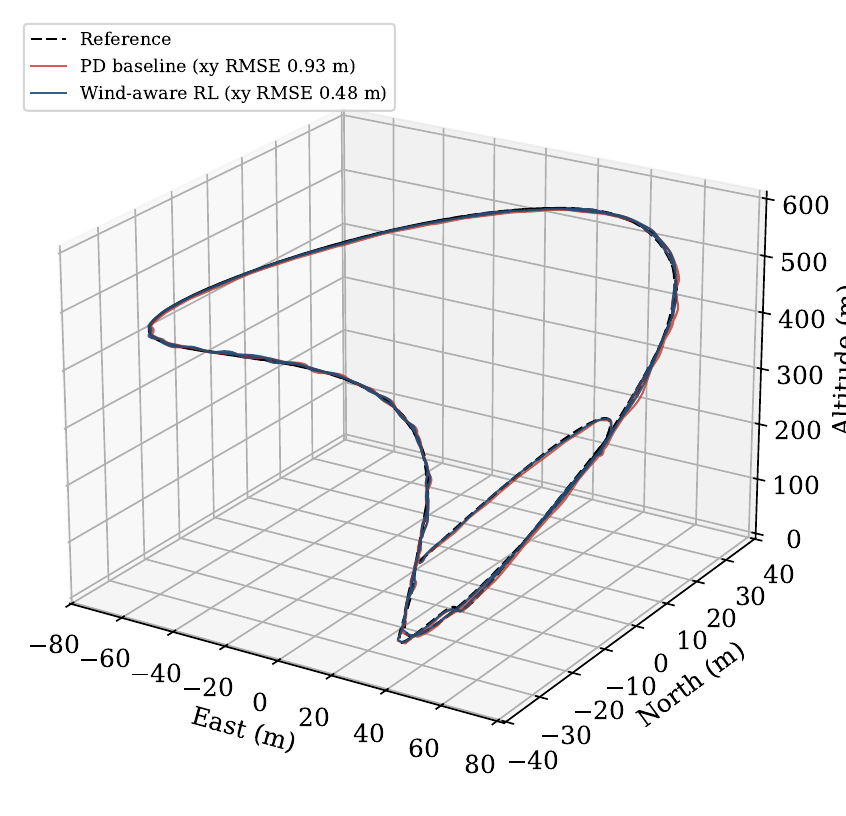}
\caption{Three-dimensional trajectory traces on the \texttt{figure8\_3d}
profile (reference wind $U_{\mathrm{mean}}=9$~m~s$^{-1}$; encountered wind
$\approx$12.7~m~s$^{-1}$ after shear amplification over the climb to
600~m), under identical replayed physics. The wind-aware RL controller
(blue) tracks the reference (dashed) more closely than the wind-blind PD
baseline (red); per-episode horizontal RMSE is 0.93~m for PD versus 0.48~m
for the wind-aware policy.}\label{fig:tracev}
\end{figure}

Per-trajectory, combined-RMSE improvement ranges from 13.2\% on
\texttt{figure8\_3d} to 46.4\% on \texttt{lawnmower\_scan}; the
figure-eight, with the most aggressive simultaneous horizontal and
vertical excitation, is the hardest to improve upon, while the scan and
climb profiles, dominated by sustained translation through the shear,
benefit most. Figure~\ref{fig:tracev} overlays traces on the
\texttt{figure8\_3d} trajectory at $U_{\mathrm{mean}} = 9$~m~s$^{-1}$,
where the horizontal perception contribution peaks.
 
\begin{table}[t]
\caption{Vertical-regime horizontal ($xy$) tracking by mean wind speed:
mean RMSE for the PD baseline, wind-blind RL, and wind-aware RL, with the
improvement over baseline and the wind-perception share of that
improvement. Each row averages 36 episodes.}\label{tab:vres}
\begin{center}
\begin{tabular}{cccccc}
\hline\hline
$U_{\mathrm{mean}}$ & PD & Wind-blind & Wind-aware & Impr. & Perception \\
(m~s$^{-1}$) & RMSE (m) & RL (m) & RL (m) & (\%) & share (\%) \\
\hline
3 & 0.576 & 0.398 & 0.409 & 28.9 & $-7.0$ \\
5 & 0.609 & 0.447 & 0.422 & 30.6 & 13.4 \\
7 & 0.734 & 0.551 & 0.439 & 40.2 & 38.0 \\
9 & 0.986 & 0.728 & 0.487 & 50.6 & 48.3 \\
\hline
Agg. & 0.726 & 0.531 & 0.440 & 39.5 & 31.9 \\
\hline
\end{tabular}
\end{center}
\end{table}
 
\subsection{Out-of-distribution winds}

\begin{figure}[!htb]
\centering
\includegraphics[width=\textwidth]{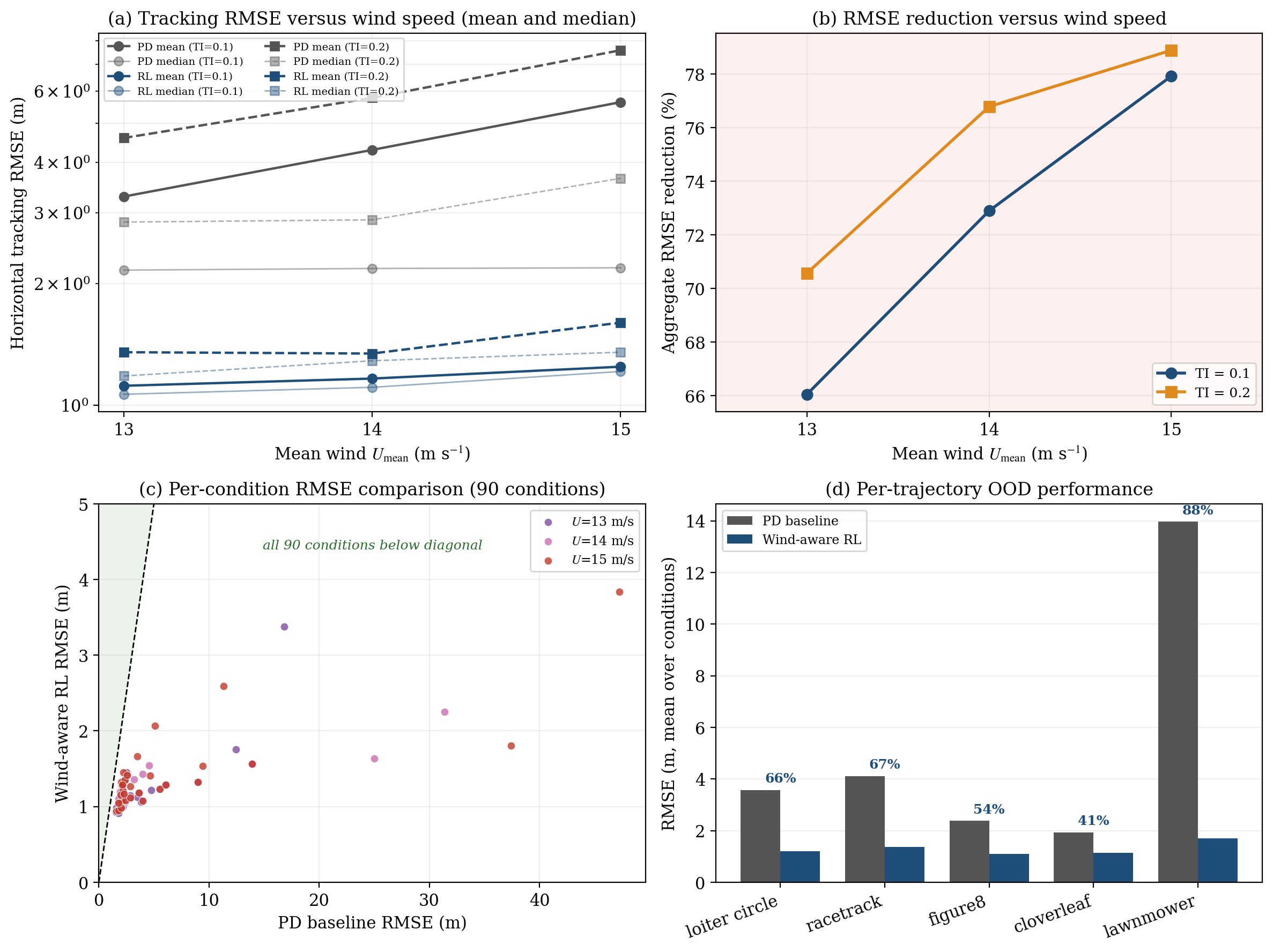}
\caption{Out-of-distribution evaluation at $U \in \{13,14,15\}$~m~s$^{-1}$,
above the $U \le 12$~m~s$^{-1}$ training range (90 conditions, 100\% win
rate, 74.9\% aggregate improvement). Top left: PD and wind-aware RL
tracking RMSE versus wind speed, mean and median; the PD mean rises far
above its median (4.40 versus 2.18~m at turbulence intensity 0.1, pooled over the three OOD wind levels), the signature of
outlier episodes in which the vehicle is lost. Top right: aggregate RMSE
reduction continuing to grow with wind through the out-of-distribution
regime, from 68.7\% at $U=13$~m~s$^{-1}$ to 78.5\% at $U=15$~m~s$^{-1}$.
Bottom left: per-condition comparison, with every condition below the
diagonal (wind-aware RL wins all 90); five PD conditions exceed a 15~m
offset while no RL condition exceeds 4~m. Bottom right: per-trajectory
mean RMSE, with the lawnmower pattern the worst-affected for PD (PD
13.96~m versus 1.70~m, an 88\% reduction).}\label{fig:ood}
\end{figure}

Evaluated unmodified at $U \in \{13, 14, 15\}$~m~s$^{-1}$, beyond its
$[3,12]$~m~s$^{-1}$ training range, the horizontal wind-aware policy
improves on the PD baseline by 74.9\% in aggregate (mean RMSE 1.30~m
versus 5.19~m) while retaining a 100\% win rate over all 90
out-of-distribution episodes. The improvement grows with wind, from
68.7\% at $U = 13$~m~s$^{-1}$ to 78.5\% at $U = 15$~m~s$^{-1}$
(Fig.~\ref{fig:ood}).

The asymmetry in failure modes is the salient result. The PD baseline
fails catastrophically: its worst-case episode RMSE reaches 47~m on the
lawnmower pattern at $U = 15$~m~s$^{-1}$, as the disturbance exceeds its
reactive authority and the vehicle is swept far off track. The wind-aware
policy, by contrast, remains bounded and controlled, with a worst-case
RMSE of 3.8~m across all out-of-distribution episodes. The failure is
outlier-driven: the PD mean diverges sharply from its median
(Fig.~\ref{fig:ood}, top left), reflecting a small number of episodes in
which the vehicle is lost entirely, and five of the ninety PD conditions
exceed a 15~m offset while every wind-aware RL condition stays below 4~m
(Fig.~\ref{fig:ood}, bottom left).

The mechanism behind this asymmetry follows from how each controller
responds to the steady wind force. The PD law is purely reactive, with no
integral action: it generates a corrective tilt only after a position
error has developed, and the steady-state error it settles to is the
displacement at which its bounded restoring force balances the wind
disturbance. Because the aerodynamic disturbance grows with the square of
the wind speed while the restoring force is capped by the commanded-tilt
limit, beyond a critical wind the disturbance can no longer be balanced
within that limit; the equilibrium error then grows without bound, and on
the upwind legs of the lawnmower pattern the vehicle is carried far off
track before the trajectory reverses, which is why the lawnmower is the
worst-affected pattern (Fig.~\ref{fig:ood}, bottom right). The wind-aware
policy instead acts in feedforward: it reads the estimated wind directly
from its observation and applies an anticipatory thrust-and-torque
correction that opposes the disturbance before an error accumulates,
rather than waiting for one to develop. This correction does not depend on
a large tracking error to activate, so it does not exhibit the same
runaway growth; even when the encountered wind exceeds anything seen in
training, the policy continues to oppose it and the error stays bounded.
Estimation accuracy is in fact highest in this strong-wind regime
(Section~\ref{sec:windest}), so the feedforward correction remains
correctly directed, and the controller degrades gracefully rather than
diverging.

The large headline improvement at these wind levels is therefore driven
substantially by baseline collapse rather than by the RL controller
performing intrinsically better than it does in distribution; it is best
read as evidence of graceful degradation, not of extrapolated mastery. An
important caveat regarding the simulation wind cap in this regime is given
in Section~\ref{sec:discussion}.

\section{Discussion and Limitations}\label{sec:discussion}

The growth of the perception share with wind speed follows from the
disturbance physics. The aerodynamic disturbance force scales
approximately as the square of the relative wind speed, so the cost of
reacting to wind, by waiting for a tracking error to develop before
correcting, grows quadratically, while the cost of small observation noise
from the wind channels is roughly constant. At low winds the noise term
can dominate, as the vertical decomposition shows with a $-7$\% perception
contribution at $U_{\mathrm{mean}} = 3$~m~s$^{-1}$. At high winds the
anticipatory benefit dominates and the wind channels carry up to half of
the total improvement. The horizontal regime adds a wrinkle at the top of
its range: the perception share peaks near $U = 8$~m~s$^{-1}$ and recedes
at $U = 12$~m~s$^{-1}$, because once the disturbance is large enough, the
gross reactive correction available to wind-blind RL accounts for the
single largest increment over the baseline, so the kinematic share grows
again even as the absolute perception benefit keeps rising. The
implication is that learned wind perception is not a uniform upgrade but a
capability concentrated in the stronger-wind conditions that define the
operational envelope of small UAS.

The vertical-axis result reinforces this picture. The $z$-axis improvement
is essentially fully kinematic, a useful negative result that localizes
the perception benefit to the horizontal plane, where the estimated
horizontal wind acts. The perception channel helps where the perceived
quantity is dynamically relevant and nowhere else, which supports the
validity of the decomposition methodology.

Several caveats bound the scope of these results. The simulation imposes a
horizontal wind-magnitude cap of 20~m~s$^{-1}$. At in-distribution wind
levels this cap is rarely active, but at the OOD levels
($U \in [13,15]$~m~s$^{-1}$) shear and turbulent gusts can drive the
instantaneous wind toward the cap, truncating the upper tail of the gust
distribution. The OOD results should therefore be read as established
under a clipped gust distribution. The qualitative finding of bounded RL
degradation versus unbounded PD failure is expected to be robust, but the
74.9\% headline figure is specific to the capped environment and is
reported with that qualification.

The decomposition quantifies the value of the estimator's output as
deployed; it does not establish how much additional improvement perfect
wind knowledge would yield. The contribution is framed through the
ablation rather than against a true-wind oracle, because the ablation
isolates exactly the information the deployed system delivers without
positing a perfect-knowledge baseline that no real vehicle could access.
The accuracy results bound the gap: in the horizontal regime, with
per-flight estimation residuals of 0.40~m~s$^{-1}$ against turbulent
fluctuations of comparable magnitude, the headroom between the deployed
estimator and an oracle is small by construction. In the vertical regime
the larger 1.56~m~s$^{-1}$ residual leaves more room, so the perception
shares reported there are a lower bound on what an ideal horizontal-wind
input could contribute.

Finally, all results are in simulation. The sensor models, aerodynamics,
and turbulence statistics, while physics-based, are not flight data, and
hardware transfer remains future work. The vertical evaluation range is
constrained by shear amplification: at representative shear exponents, winds
encountered near 600~m altitude substantially exceed $U_{\mathrm{mean}}$,
so $U_{\mathrm{mean}}$ must be capped to keep the encountered wind within
the estimator's training distribution. The estimator and policy are also
trained on the trajectory families of Section~\ref{sec:simenv}, so
generalization to qualitatively different maneuvers such as aggressive
aerobatics is untested.

\section{Conclusion}\label{sec:conclusion}

This work presented a two-stage pipeline for wind-aware flight of a small
quadrotor in simulated atmospheric turbulence: a frozen attention-GRU
estimator that recovers the ambient wind vector from onboard kinematics at
accuracies near the turbulence-imposed floor, and a PPO controller that
exploits the estimate as a feedforward correction to a classical
proportional-derivative law. Relative to the wind-blind baseline, the
wind-aware controller reduces horizontal tracking error by 48\% and
vertical horizontal-axis error by 39.5\%, winning every in-distribution
evaluation episode. A three-way ablation decomposed these gains into
kinematic and wind-perception components and showed that the perception
share increases with wind speed, from negligible or slightly negative in
light winds to roughly half the total benefit in strong winds, mirroring
the quadratic scaling of aerodynamic disturbance forces. The same
decomposition localized the perception benefit to the horizontal plane,
where the estimated wind acts, and found it absent on the vertical axis,
internal evidence that the channel operates through the physically correct
mechanism rather than as a generic capacity gain. On out-of-distribution
winds the learned controller degraded gracefully while the classical
baseline diverged, a contrast that follows directly from the difference
between anticipatory feedforward and purely reactive control.

The central finding of this is that learned wind perception is a high-leverage,
sensor-free addition to small-UAS autonomy whose value concentrates in the
strong-wind conditions where reliable flight is hardest. Future work
includes extending the vertical evaluation to higher reference winds with
reduced shear exponents, treating vertical wind-component estimation and
tracking as a research thread in its own right, and ultimately validating
the pipeline in hardware flight.

\FloatBarrier

\bibliography{references}

\end{document}